\documentclass{article}
\usepackage{preprint,times}

\usepackage[colorlinks=false,breaklinks=true,bookmarks=false,citecolor=green]{hyperref}
\usepackage{url}
\usepackage{amsmath,amssymb}
\usepackage{booktabs}
\usepackage{graphicx}
\usepackage{enumitem}
\usepackage{multirow}
\usepackage{tabularx}
\usepackage{xcolor}

\usepackage{xspace}
\newcommand{\NICKNAME}{\textsc{HiP3R}}
\newcommand{\method}{\NICKNAME\xspace}

\newcommand{\ate}{ATE}
\newcommand{\rper}{RPE\textsubscript{rot}}
\newcommand{\rpet}{RPE\textsubscript{trans}}
\newcommand{\stgv}[2]{$#1_{\scriptscriptstyle(#2)}$}

\title{\method{}: Hierarchical Pose Graphs for Long-Horizon 3D Reconstruction}

\author{%
\hspace{-2.5mm} Zunnan Xu$^{1}$\quad Shangchen Zhou$^{2}$\quad Yihang Luo$^{1}$\quad Chen Change Loy$^{1,3,}\thanks{Corresponding author.}$
\vspace{0.7mm}\\
\hspace{-2.5mm} $^{1}$S-Lab, Nanyang Technological University\quad $^{2}$Imperial College London\quad $^{3}$ACE Robotics
\vspace{0.7mm}\\
\hspace{-1.5mm}{\tt\small \url{https://kkakkkka.github.io/HiP3R}}
}

\begin{document}

\maketitle

\begin{abstract}
Feed-forward models accurately estimate camera motion and 3D geometry from short video windows.
However, computational and memory costs prevent direct inference on sequences containing thousands of frames.
Long videos must therefore be processed as overlapping chunks.
This division produces independent local reconstructions with inconsistent coordinate frames and duplicate estimates of shared cameras.
Achieving global consistency requires propagating constraints from adjacent overlaps and distant revisits across the sequence.
This requirement naturally motivates pose graph optimization.
However, the choice of graph level creates a trade-off.
\textit{Chunk-level} graphs efficiently resolve scale drift and close large loops.
Yet, their rigid transformations cannot correct internal errors or conflicting frame predictions.
Conversely, \textit{frame-level} optimization offers local flexibility, but scaling it globally is computationally expensive and can break metric scale.
\method{} resolves this trade-off using a hierarchical pose graph.
\method{} first optimizes a robust chunk-level $\mathrm{Sim}(3)$ graph to establish the global layout and scale.
It then holds this layout fixed and estimates one shared $\mathrm{SE}(3)$ camera per frame, turning overlapping window predictions into a single trajectory.
Experiments across multiple benchmarks demonstrate the effectiveness of \method{} in improving camera trajectories and 3D reconstruction.
Our method applies to both chunk-based and streaming backends without additional training or fine-tuning.
\end{abstract}

\section{Introduction}
\label{sec:intro}

Recovering globally consistent 3D geometry and camera trajectories from video is fundamental to computer vision.
This capability supports autonomous navigation, robotic mapping, and spatial artificial intelligence.
Recent 3D foundation models~\citep{wang2024dust3r,wang2025vggt,wang2026vggtomega} have significantly advanced visual geometry.
By jointly predicting camera poses and dense pointmaps in a single forward pass, they bypass classical handcrafted matching.
These models achieve high accuracy on short video clips without per-scene calibration or optimization.
However, the quadratic complexity of visual attention limits them to short temporal windows.
Directly scaling transformers to extended multi-kilometer sequences remains too expensive.

To process long videos, modern systems adopt a divide-and-conquer strategy~\citep{deng2025vggtlong}.
Long sequences are divided into overlapping video chunks, which are processed independently and stitched together.
While chunking bounds GPU memory, sequential alignment introduces severe computational and numerical bottlenecks.
Dense overlapping points share high spatial redundancy, making all-pixel alignment computationally expensive.
Furthermore, pairwise alignment errors compound along the sequential chain, causing severe scale and orientation drift.
Loop closures are essential to eliminate cumulative drift and enforce global topological consistency.
However, when the chains are not yet aligned, applying multiple long-range loop constraints at once can make the non-convex optimization unstable. The solver may diverge or converge to poor local minima.

\begin{figure*}[t]
\centering
\includegraphics[width=\textwidth]{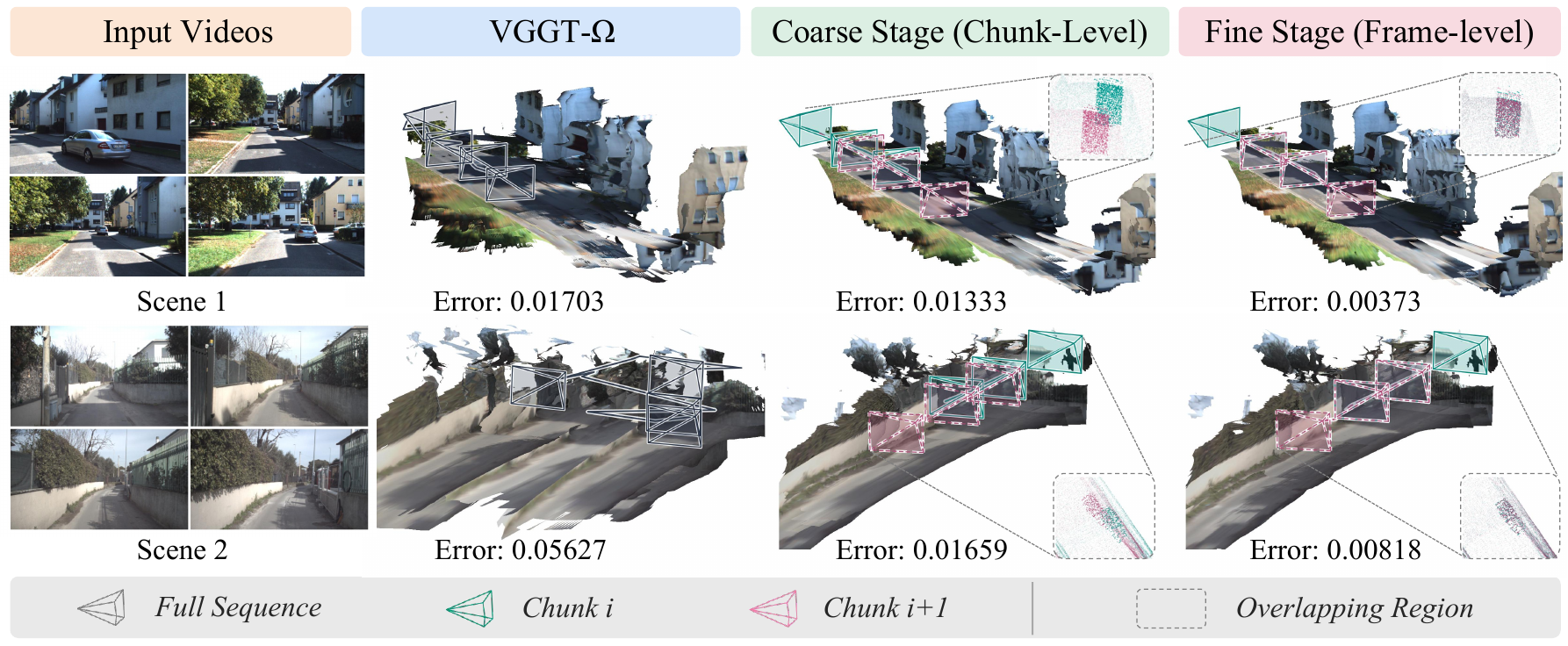}
\caption{Overlapping predictions recover local geometry in inconsistent frames. Coarse layout optimization corrects global placement and scale, but keeps each chunk's internal cameras. Shared-camera trajectory refinement estimates one trajectory from predictions and within-chunk motion.}
\label{fig:teaser}
\end{figure*}

As shown in Figure~\ref{fig:teaser}, even after global chunk placement is resolved, a conflict remains at the frame level.
Existing alignment methods~\citep{deng2025vggtlong,zhang2026vggtalign,xiong2026vggtmotion} optimize similarity transformations per chunk.
However, a single similarity transform preserves internal camera configurations up to scale.
Consequently, it cannot resolve cross-window camera-pose inconsistency.
Overlapping chunks produce multiple, conflicting pose estimates of the same frame under differing visual contexts.
A chunk is an inference unit, whereas the true trajectory contains only one camera pose per frame.
Arbitrarily selecting one prediction discards multi-view evidence and causes velocity jumps across chunk boundaries.
Conversely, full bundle adjustment is too expensive, while relative motion alone drifts from the recovered global layout.

To overcome these challenges, we introduce \method{}, a {Hi}erarchical {P}ose-graph optimization method for long-horizon {3}D {R}econstruction.
\method{} decomposes reconstruction into \textit{coarse chunk} layout recovery and \textit{fine frame} trajectory refinement.
The coarse stage optimizes a \textit{chunk-level} $\mathrm{Sim}(3)$ pose graph over adjacent overlaps and distant loop closures.
This stage establishes a globally consistent layout and eliminates accumulated scale drift.
Holding this coarse layout fixed, the fine stage optimizes a \textit{frame-level} shared-camera \(\mathrm{SE}(3)\) pose graph, representing each physical frame with a single camera variable to resolve cross-window pose inconsistency.
Placement priors keep that camera on the recovered layout, while within-chunk motion factors resolve conflicting predictions from overlapping windows.
We evaluate \method{} across seven public benchmarks spanning diverse indoor and outdoor environments.
Under matched backends and windows, \method{} improves long-horizon trajectory accuracy over loop-enabled VGGT-Long.
Controlled ablations show that bounded point sampling reduces graph construction and optimization time with little change in accuracy.
Additional evaluations show that the method also works with streaming backends such as LingBot-MAP.
We summarize our main contributions as follows:
\begin{itemize}[leftmargin=*, topsep=0pt, itemsep=2pt]
\item We show that chunk-level similarity alignment alone cannot ensure cross-window camera-pose consistency, as overlapping windows may yield conflicting estimates for the same frame.
\item We develop a hierarchical pose graph that first recovers a globally consistent chunk layout and scale, then estimates one shared $\mathrm{SE}(3)$ camera pose for each frame.
\item We introduce progressive loop closure and bounded point sampling to achieve stable and scalable optimization on long sequences without retraining.
\end{itemize}

\section{Related Work}
\label{sec:related}

\noindent\textbf{Feed-forward 3D vision models.}
Feed-forward 3D models have emerged as an alternative to classical Structure-from-Motion (SfM).
Pioneering architectures like DUSt3R~\citep{wang2024dust3r} and MASt3R~\citep{leroy2024mast3r} directly regress dense pointmaps from uncalibrated image pairs.
Recent multi-view transformers such as VGGT~\citep{wang2025vggt} and VGGT-Omega~\citep{wang2026vggtomega} extend feed-forward predictions across broader view collections.
Subsequent models explore permutation invariance~\citep{wang2025pi3}, sparse attention~\citep{wang2026blocksparse}, and learned scene actions~\citep{wang2026amb3r}.
These models achieve high local geometric accuracy without per-scene numerical optimization.
However, quadratic attention restricts them to short windows, so long sequences yield overlapping local reconstructions.

\noindent\textbf{Streaming and recurrent video geometry.}
To extend temporal context beyond static windows, another line of research develops streaming and recurrent architectures.
Early methods such as Spann3R~\citep{wang2024spann3r} and CUT3R~\citep{wang2025cut3r} maintain persistent spatial memory states.
Recent methods like StreamVGGT~\citep{zhuo2026streamvggt}, STream3R~\citep{lan2026stream3r}, and WinT3R~\citep{li2026wint3r} process video frames causally with temporal key-value caching.
LingBot-MAP~\citep{chen2026lingbotmap} structures geometric context across anchor frames, local windows, and trajectory memory.
Other approaches explore test-time training, memory compression, and feature retrieval~\citep{chen2026ttt3r,zhang2026loger,xie2026scal3r,jin2026zipmap,liu2026streamcachevggt,zou2026retrievevggt}.
Causal streaming still accumulates drift on long trajectories.
As post-processing, \method{} adds global loops and per-frame pose consistency without retraining.

\noindent\textbf{Multi-window fusion and Structure-from-Motion.}
Scaling feed-forward predictions to extended trajectories connects to classical global Structure-from-Motion and pose graph optimization.
Chunk-based methods like VGGT-Long~\citep{deng2025vggtlong}, VGGT-Align~\citep{zhang2026vggtalign}, and VGGT-Motion~\citep{xiong2026vggtmotion} align adjacent chunks via pairwise $\mathrm{Sim}(3)$ alignment.
A rigid $\mathrm{Sim}(3)$ per chunk preserves internal cameras up to scale and cannot resolve overlapping-frame disagreements.
GlueMap~\citep{pan2026gluemap} couples feed-forward star reconstructions with global motion averaging and bundle adjustment over virtual tracks.
\method{} resolves overlapping camera predictions through a shared frame-level pose graph after chunk alignment, without optimizing 3D tracks.

\begin{figure*}[t]
\centering
\includegraphics[width=\textwidth]{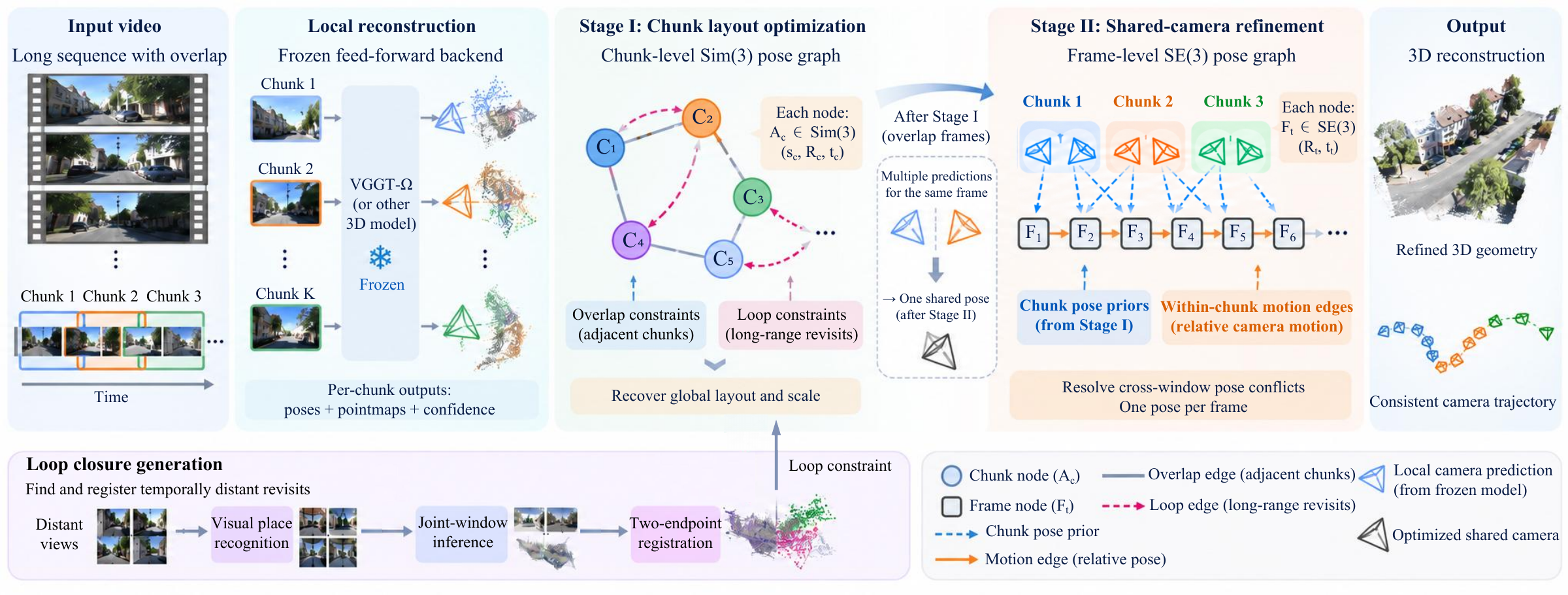}
\vspace{-2mm}
\caption{Overview of \method{}. Frozen backends predict overlapping chunks. Stage I builds a chunk-level $\mathrm{Sim}(3)$ graph from overlap and loop constraints to recover layout and scale. Stage II estimates one shared $\mathrm{SE}(3)$ camera per frame under that layout.}
\label{fig:framework}
\end{figure*}

\section{Method}
\label{sec:method}

\subsection{Problem Formulation}
\label{sec:problem_formulation}

Let $\mathcal{I} = \{I_t\}_{t=1}^{T}$ denote an input video sequence of $T$ frames.
We partition $\mathcal{I}$ into $K$ overlapping chunks $\{C_k\}_{k=1}^{K}$ of at most $C$ frames with overlap $O$.
Chunk starts advance by $C-O$ frames; the final chunk may be shorter than $C$.
Each chunk is processed independently by a feed-forward 3D vision model~\citep{wang2026vggtomega}.
For chunk $C_k$, the backend predicts dense pointmaps $X_{k,i}$, confidence maps $\gamma_{k,i}$, and camera-to-chunk poses $P_{k,i}$.
Here $i \in \{1,\dots,|C_k|\}$ indexes local frames within the chunk.
Let $g(k,i) \in \{1,\dots,T\}$ map local frame $i$ in chunk $C_k$ to its global video index.
The goal is to recover a single camera trajectory $\mathcal{F} = \{F_t \in \mathrm{SE}(3)\}_{t=1}^{T}$ together with consistent 3D pointmaps.
Figure~\ref{fig:framework} gives an overview.

\noindent \textbf{Decoupled optimization.} Directly optimizing all camera and point variables across thousands of frames is expensive.
Unconstrained bundle adjustment loses the efficiency of feed-forward models.
\method{} uses a two-level hierarchical pose graph.
Stage~I solves a chunk-level $\mathrm{Sim}(3)$ pose graph for a globally aligned layout (Section~\ref{sec:chunks}).
Stage~II then optimizes a frame-level $\mathrm{SE}(3)$ pose graph with one shared camera variable per frame while holding the layout fixed (Section~\ref{sec:unique}).

\subsection{Coarse Layout Optimization with Progressive Loop Scheduling}
\label{sec:chunks}

We formulate global layout recovery as optimization over a similarity pose graph $\mathcal{G}_c = (\mathcal{V}_c, \mathcal{E}_c)$.
Each node $k \in \mathcal{V}_c$ represents a chunk $C_k$ and is associated with a similarity transformation $A_k = (s_k, R_k, t_k) \in \mathrm{Sim}(3)$.
This transformation maps local chunk coordinates to a common world coordinate system as $A_k(x) = s_k R_k x + t_k$.
An edge $(i,j) \in \mathcal{E}_c$ represents a measured relative transformation $\widehat Z_{ij}$ from the coordinate system of chunk $i$ to that of chunk $j$.
The corresponding predicted relative transformation is $A_j^{-1} A_i$.
Overlap edges connect consecutive chunks, while loop edges connect retrieved chunk pairs that observe the same physical scene.
Relative constraints determine scale only up to a global factor, so we fix $A_1$ as the world frame and scale reference.
Scale errors in independent chunk predictions accumulate on long trajectories, so we optimize the remaining $s_k$ jointly with rotation and translation.

Dense overlapping 3D points are spatially redundant, so all-pixel alignment is unnecessary.
For adjacent chunks, identical pixels in overlapping frames yield paired 3D predictions.
We estimate relative similarities with confidence-weighted iteratively reweighted least squares (IRLS) rather than unweighted Umeyama alignment on all dense pixels~\citep{deng2025vggtlong}.
Each iteration runs a weighted Umeyama alignment~\citep{umeyama1991least} and updates Huber weights from spatial residuals.
We bound alignment cost by limiting the number of point correspondences used in each fit.

Visual place recognition proposes candidate loop closures across temporal separations~\citep{izquierdo2024salad}.
For each selected pair of revisiting frames, we combine temporal neighborhoods from both endpoints into one backend input.
The resulting joint reconstruction connects the two visits in a common local coordinate system.
We align the predicted geometry at each endpoint to the corresponding original chunk, obtaining transformations $B_i$ and $B_j$ from the joint reconstruction to chunks $i$ and $j$, respectively.
The relative loop measurement is $\widehat Z_{ij} = B_j B_i^{-1}$.
We assign each loop constraint a bounded weight $w_{ij}$ based on the registration residuals at its two endpoints, reducing the influence of poorly aligned loops.
Appendix~\ref{app:protocol} specifies loop retrieval and window settings, point sampling, pairwise alignment, and loop weighting.

Combining sequential overlap and loop constraints, we define the objective over active edges $\mathcal E_c$.
Let $E^c_{ij} = \widehat Z_{ij}^{-1} A_j^{-1} A_i$ denote the relative similarity error.
We define the residual vector as
\begin{equation}
    r^c_{ij}=
    \big(\mathrm{Log}_{\mathrm{SO}(3)}R(E^c_{ij}),\,
    \tau(E^c_{ij}),\,\log s(E^c_{ij})\big)\in\mathbb R^7,
    \label{eq:coarse_residual}
\end{equation}
where $\tau$ extracts Euclidean translation error, and $\log s$ evaluates log-scale error.
Because feed-forward backends predict geometry without analytical pose covariances,
$w_{ij}$ provides a data-driven confidence weight per edge.
The coarse layout objective minimizes the robust loss
\begin{equation}
    A^\star=\arg\min_A
    \sum_{(i,j)\in\mathcal E_c}\sum_{\ell=1}^{7}
    \rho_\delta\!\left(\sqrt{w_{ij}}\,r^c_{ij,\ell}\right),
    \label{eq:coarse}
\end{equation}
where $\rho_\delta$ denotes the Huber penalty, applied coordinate-wise to suppress large outliers.

Activating all distant loops on an unaligned chain can make the non-convex optimization unstable.
We initialize chunk placements from sequential alignments, then introduce loop constraints in order of increasing temporal span.
Adjacent edges remain active throughout, and each stage starts from the solution of the preceding stage.
When most loop edges have short temporal spans, additional stages introduce the longer-span constraints more gradually.

We solve Eq.~\eqref{eq:coarse} with bounded trust-region reflective least squares, limiting log-scale changes relative to initialization.
Each edge residual depends only on the transformations of the two connected chunks, yielding a sparse Jacobian.
We compute the corresponding Jacobian blocks using automatic differentiation.
Appendix~\ref{app:protocol} provides the parameterization, scheduling rules, and restart settings.

\subsection{Shared-Camera Trajectory Refinement}
\label{sec:unique}

Given the recovered layout $A^\star$, overlapping frames still have multiple conflicting pose estimates.
Selecting one prediction discards alternative observations and causes velocity jumps across chunk boundaries.
We therefore introduce a shared camera-to-world variable $F_t \in \mathrm{SE}(3)$ for each frame $t$.
We represent each camera pose $P_{k,i}$ by its rotation $R_{k,i} \in \mathrm{SO}(3)$ and camera center $c_{k,i} \in \mathbb{R}^3$, writing $P_{k,i} = [R_{k,i} \mid c_{k,i}]$.
Its placement $Q_{k,i}$ in the world frame is
\begin{equation}
    R(Q_{k,i})=R_kR_{k,i},\qquad
    c(Q_{k,i})=s_kR_kc_{k,i}+t_k,
    \label{eq:place}
\end{equation}
with $A_k = A_k^\star$.
The chunk scale applies to camera centers.
The shared pose is rigid with no scale variable.
For each shared camera, we use one coarse-placed prediction as its initial pose.
The placement priors retain the predictions from every chunk that observes that frame.

As shown in Figure~\ref{fig:framework}, the camera graph uses two types of geometric factors.
For consecutive frames $i,j$ retained as graph nodes within chunk $k$, define the relative motion measurement $M_{k,i\rightarrow j} = Q_{k,j}^{-1} Q_{k,i}$.
With $E^f = M_{k,i\rightarrow j}^{-1} F_{g(k,j)}^{-1} F_{g(k,i)}$, the relative motion edge residual is
\begin{equation}
    r^f_{k,i,j}=\big(\mathrm{Log}_{\mathrm{SO}(3)}R(E^f),\,\tau(E^f)\big)\in\mathbb R^6.
    \label{eq:fine}
\end{equation}
Here $\tau$ extracts Euclidean translation, and the rotation residual uses the standard rotation logarithm~\citep{sola2018micro}.
Each placed prediction also provides a unary chunk pose prior on its shared camera, with $E^p = Q_{k,i}^{-1} F_{g(k,i)}$ and
\begin{equation}
    r^p_{k,i}=\big(\mathrm{Log}_{\mathrm{SO}(3)}R(E^p),\,\tau(E^p)\big)\in\mathbb R^6.
    \label{eq:prior}
\end{equation}
When a physical frame appears in multiple overlapping chunks, their respective chunk pose priors jointly constrain the same shared variable $F_t$. Relative motion edges connected to this node join local chunk trajectories across chunk boundaries. The fine objective therefore combines both relative motion constraints and placement priors,
\begin{equation}
\begin{split}
    F^\star=\arg\min_{F} &
    \sum_{(k,i,j)\in\mathcal E_f}\sum_{\ell=1}^{6}
    \rho_\delta\!\left(\sqrt{\lambda_f}\,r^f_{k,i,j,\ell}(A^\star,F)\right)
    +\sum_{(k,i)\in\mathcal E_p}\sum_{\ell=1}^{6}
    \rho_\delta\!\left(\sqrt{\lambda_p}\,r^p_{k,i,\ell}(A^\star,F)\right).
\end{split}
    \label{eq:fine_objective}
\end{equation}
Motion constraints preserve relative poses within each chunk, while placement priors keep the refined trajectory close to the recovered global layout.
Initialization and solver settings are given in Appendix~\ref{app:protocol}, and closed-form Jacobian blocks in Appendix~\ref{app:jacobian}.

For long sequences, we optimize a keyframe subgraph for efficiency.
Let $Q_t$ denote the coarse-placed pose used to initialize frame $t$, and define its correction as $\Delta_t = F_t Q_t^{-1}$.
We interpolate corrections between keyframes and set $F_t = \Delta_t Q_t$.
This recovers a trajectory for all frames while retaining their local pose predictions as the baseline.
To update the geometry, we first place each pointmap in world coordinates using $A_k^\star$, then apply the rigid correction $F_{g(k,i)} Q_{k,i}^{-1}$.
The corrected pointmaps follow the shared camera poses while preserving the predicted depth and recovered chunk scale.
Appendix~\ref{app:keyframe} describes keyframe selection and correction interpolation.

\section{Experiments}
\label{sec:experiments}

We evaluate trajectory accuracy across seven public benchmarks comprising $61$ diverse sequences.
The benchmark spans indoor room-scale environments, including 7-Scenes~\citep{shotton2013scene} and Neural RGB-D~\citep{azinovic2022neural}.
It also includes high-precision multi-scale scenes from ETH3D~\citep{schops2017eth3d} and Tanks and Temples~\citep{knapitsch2017tanks}.
Finally, it includes extended outdoor driving and campus traversals from Oxford Spires~\citep{tao2024spires}, KITTI~\citep{geiger2012kitti}, and VBR~\citep{brizi2024vbr}.
Dense surface geometry is evaluated on ETH3D, Neural RGB-D, and 7-Scenes, which provide dense 3D ground truth.
Appendix~\ref{app:datasets} details dataset compositions and sequence splits.
Baselines in Table~\ref{tab:external} use released checkpoints and recommended window policies on identical sequences.
VGGT-Long and VGGT-Align are evaluated from public implementations, while VGGT-Motion~\citep{xiong2026vggtmotion} is not publicly available.

We evaluate trajectory accuracy using \ate{}, \rper{}, and \rpet{} (lower is better).
\ate{} is translation RMSE after a per-sequence $\mathrm{Sim}(3)$ alignment that corrects rotation, translation, and scale.
Relative errors use all consecutive frame pairs ($\Delta{=}1$ frame).
Rotation RPE is in degrees, and translation RPE applies the same $\mathrm{Sim}(3)$ correction.
Pairwise pose AUC (higher is better) evaluates all $i<j$ camera pairs using the maximum of relative rotation and translation-direction angular errors at thresholds of $5^{\circ}$, $15^{\circ}$, and $30^{\circ}$.
Reconstructed 3D pointmaps are evaluated using the mean of accuracy and completeness (Chamfer, meters) after Umeyama and ICP.
F1 uses a hit distance of $0.25$\,m on ETH3D and $0.05$\,m on Neural RGB-D and 7-Scenes.
Appendix~\ref{app:e2e} provides  per-dataset geometric breakdowns.

We apply \method{} to the camera poses and dense geometry produced by {VGGT-Omega}~\citep{wang2026vggtomega} and {LingBot-MAP}~\citep{chen2026lingbotmap}.
Both backends use the same graph formulation without fine-tuning.
Unless otherwise stated, we use VGGT-Omega 1B with a chunk size of $60$ frames and an overlap size of $30$ frames.
LingBot-MAP uses windowed inference with a chunk size of $120$ frames and an overlap size of $32$ frames.

We use the same refinement weights across backends.
For short sequences, the fine stage optimizes all camera poses; for long sequences, it optimizes a keyframe subgraph and interpolates pose corrections for the remaining frames.
Appendix details the keyframe selection rules and solver settings.

\subsection{Benchmark Evaluation across Public Datasets}
\label{sec:benchmark_eval}

Table~\ref{tab:external} reports tracking metrics macro-averaged across the seven datasets.
With VGGT-Omega, \method{} reduces \ate{} by $74\%$, from $18.555$ to $4.906$ m, \rper{} from $1.820^\circ$ to $0.590^\circ$, and \rpet{} by $94\%$, from $0.9878$ to $0.0622$ m.
On long traversals, sequential stitching accumulates scale and orientation drift.
Progressive loop scheduling in the coarse layout stage (Section~\ref{sec:chunks}) corrects global drift and drives much of the \ate{} reduction in Table~\ref{tab:external}.
Shared-camera refinement (Section~\ref{sec:unique}) further improves relative pose accuracy while preserving the layout.
On compact indoor benchmarks such as 7-Scenes and Neural RGB-D, native VGGT-Omega already achieves \ate{} of $0.037$ m and $0.019$ m, leaving limited room for further global correction.
With LingBot-MAP, \method{} reduces macro \ate{} by $52\%$, from $8.288$ to $4.017$ m, without retraining.

\begin{table}[ht]
\centering
\vspace{-2mm}
\caption{Trajectory evaluation across indoor and outdoor benchmarks.
Values are dataset-macro averages of \ate{} (m), \rper{} ($^\circ$), and \rpet{} (m).
Bold marks the best result.}
\vspace{1mm}
\label{tab:external}
\small
\setlength{\tabcolsep}{4.5pt}
\begin{tabular}{lrrr}
\toprule
\textbf{Method} &
\textbf{Mean \ate{}} $\downarrow$ &
\textbf{Mean \rper{}} $\downarrow$ &
\textbf{Mean \rpet{}} $\downarrow$ \\
\midrule
CUT3R~\citep{wang2025cut3r} & 44.894 & 4.469 & 1.4600 \\
TTT3R~\citep{chen2026ttt3r} & 23.775 & 5.077 & 0.8077 \\
LoGeR~\citep{zhang2026loger} & 40.691 & 2.086 & 1.1035 \\
VGGT~\citep{wang2025vggt} & 20.914 & 3.159 & 1.6396 \\
ZipMap~\citep{jin2026zipmap} & 28.230 & 2.792 & 0.9454 \\
LingBot-MAP~\citep{chen2026lingbotmap} & 8.288 & 1.818 & 1.4152 \\
VGGT-Long~\citep{deng2025vggtlong} & 22.713 & 1.449 & 0.2537 \\
VGGT-Align~\citep{zhang2026vggtalign}& 25.584 & 1.714 & 2.3608 \\
VGGT-Omega~\citep{wang2026vggtomega} & 18.555 & 1.820 & 0.9878 \\
$\pi^3$~\citep{wang2025pi3} & 24.029 & 1.592 & 0.3614 \\
GlueMap~\citep{pan2026gluemap} & 45.281 & 1.561 & 1.9132 \\
Scal3R~\citep{xie2026scal3r} & 27.612 & 1.376 & 0.1703 \\
\midrule
\textbf{\method{}} (LingBot-MAP) & \textbf{4.017} & 1.055 & 0.1824 \\
\textbf{\method{}} (VGGT-Omega) & 4.906 & \textbf{0.590} & \textbf{0.0622} \\
\bottomrule
\end{tabular}
\end{table}

\subsection{Controlled Long-Horizon Reconstruction}
\label{sec:kitti_vbr}

We compare \method{} with two VGGT-Long~\citep{deng2025vggtlong} variants, with and without loop closures.
All three use the same VGGT-Omega backend and the default chunk and overlap sizes specified above.
Table~\ref{tab:long_cmp} reports \ate{} on KITTI, VBR, and Oxford and the mean over these three datasets.
For reference, it also includes the released VGGT-Long and VGGT-Align~\citep{zhang2026vggtalign} results using VGGT predictions.

\method{} achieves lower \ate{} than both VGGT-Long variants using VGGT-Omega on all three datasets.
Loop closures reduce VGGT-Long \ate{} from $110.3$ to $10.19$ m on KITTI and from $85.50$ to $41.27$ m on VBR.
\method{} further reduces these errors to $9.730$ and $22.44$ m, nearly halving the remaining VBR error.
VGGT-Long applies loops as rigid $\mathrm{Sim}(3)$ transforms to entire chunks.
\method{} instead jointly optimizes a sparse global layout graph with progressive loop scheduling.
Appendix~\ref{app:layout_controls} shows that the complete solver further improves accuracy and convergence efficiency, while Appendix~\ref{app:long_time} reports stage-wise runtimes.

\begin{table}[ht]
\centering
\caption{Trajectory comparison on KITTI, VBR, and Oxford, alongside the long-horizon mean.}
\vspace{1mm}
\label{tab:long_cmp}
\small
\setlength{\tabcolsep}{3.2pt}
\begin{tabular}{ll cccc}
\toprule
& & \multicolumn{4}{c}{\textbf{\ate{}} $\downarrow$} \\
\cmidrule(lr){3-6}
\textbf{Backend} & \textbf{Method} & \textbf{Oxford} & \textbf{KITTI} & \textbf{VBR} & \textbf{Mean} \\
\midrule
\multirow{4}{*}{{VGGT-Omega}}
& Baseline & $9.861$ & $77.42$ & $41.91$ & $43.06$ \\
& VGGT-Long (w/o loops) & $1.777$ & $110.3$ & $85.50$ & $65.86$ \\
& VGGT-Long (w/ loops) & $1.665$ & $10.19$ & $41.27$ & $17.71$ \\
& $+$ \method{} (Ours) & \textbf{1.579} & \textbf{9.730} & \textbf{22.44} & \textbf{11.25} \\
\midrule
\multirow{2}{*}{{VGGT}}
& VGGT-Long (released) & $9.451$ & $48.84$ & $99.83$ & $52.71$ \\
& VGGT-Align & $8.567$ & $69.80$ & $99.87$ & $59.41$ \\
\bottomrule
\end{tabular}
\end{table}

\begin{figure*}[t]
\centering
\includegraphics[width=\textwidth]{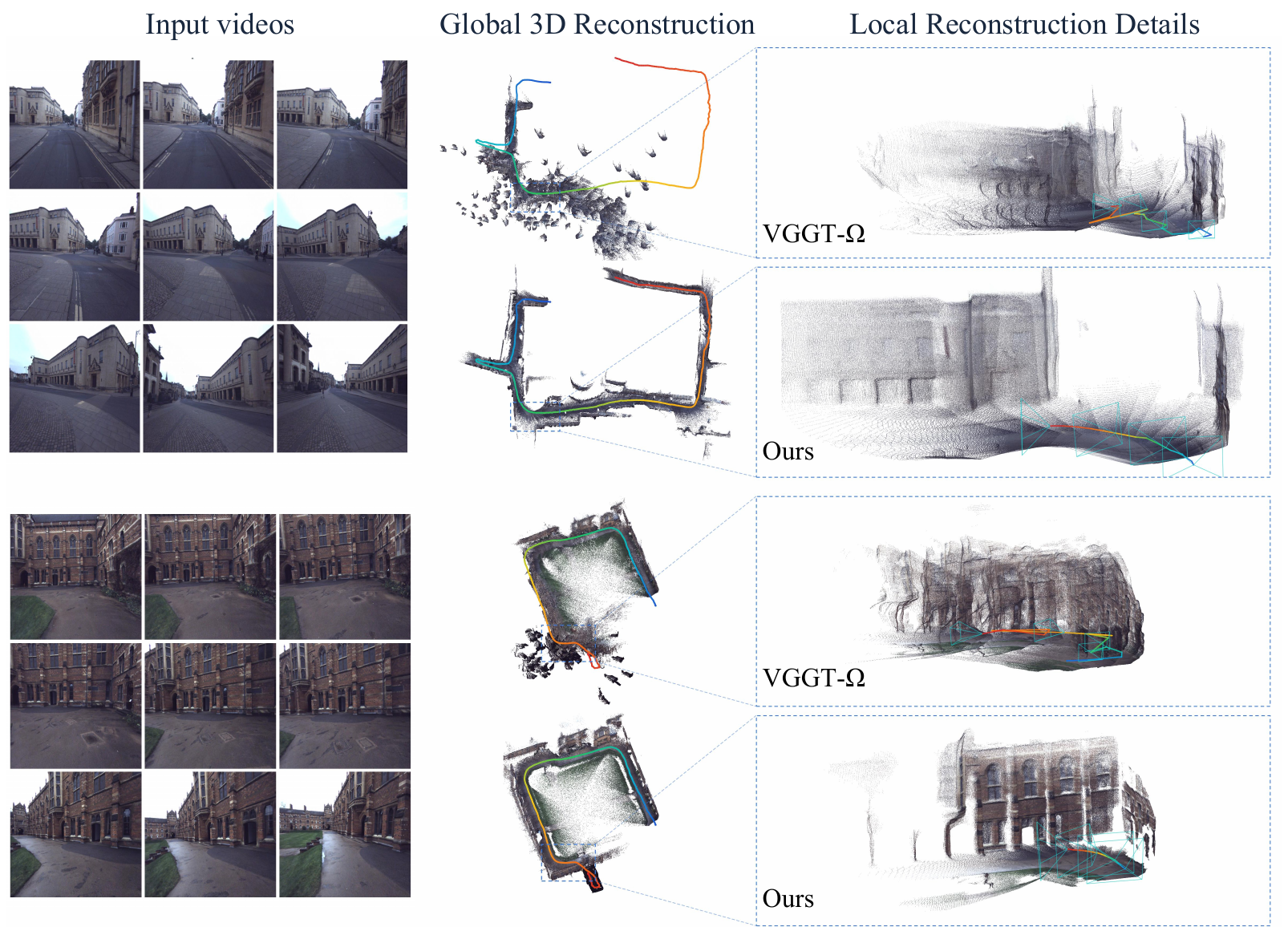}
\vspace{-2mm}
\caption{Qualitative comparison between VGGT-Omega chunk placement and \method{}.}
\label{fig:recon_comparison}
\vspace{-3mm}
\end{figure*}

\subsection{End-to-End Pose Accuracy and 3D Geometry}
\label{sec:e2e}

Inconsistent poses place walls and corridors at conflicting locations, causing duplicated surfaces and ghosting.
Table~\ref{tab:e2e_omega} reports pose AUC and surface metrics on both backends.
With VGGT-Omega, \method{} reduces Chamfer by $64\%$, from $0.0759$ to $0.0272$ m, and raises F1 from $65.92$ to $91.26$.
The stage-wise comparison in Appendix~\ref{app:step_attribution}, Table~\ref{tab:step_geometry}, shows that coarse alignment accounts for most of the Chamfer reduction, with improvements from shared-camera refinement on both backends.

\subsection{Qualitative Dense 3D Reconstruction}
\label{sec:qualitative}

\begin{figure*}[t]
\centering
\includegraphics[width=\textwidth]{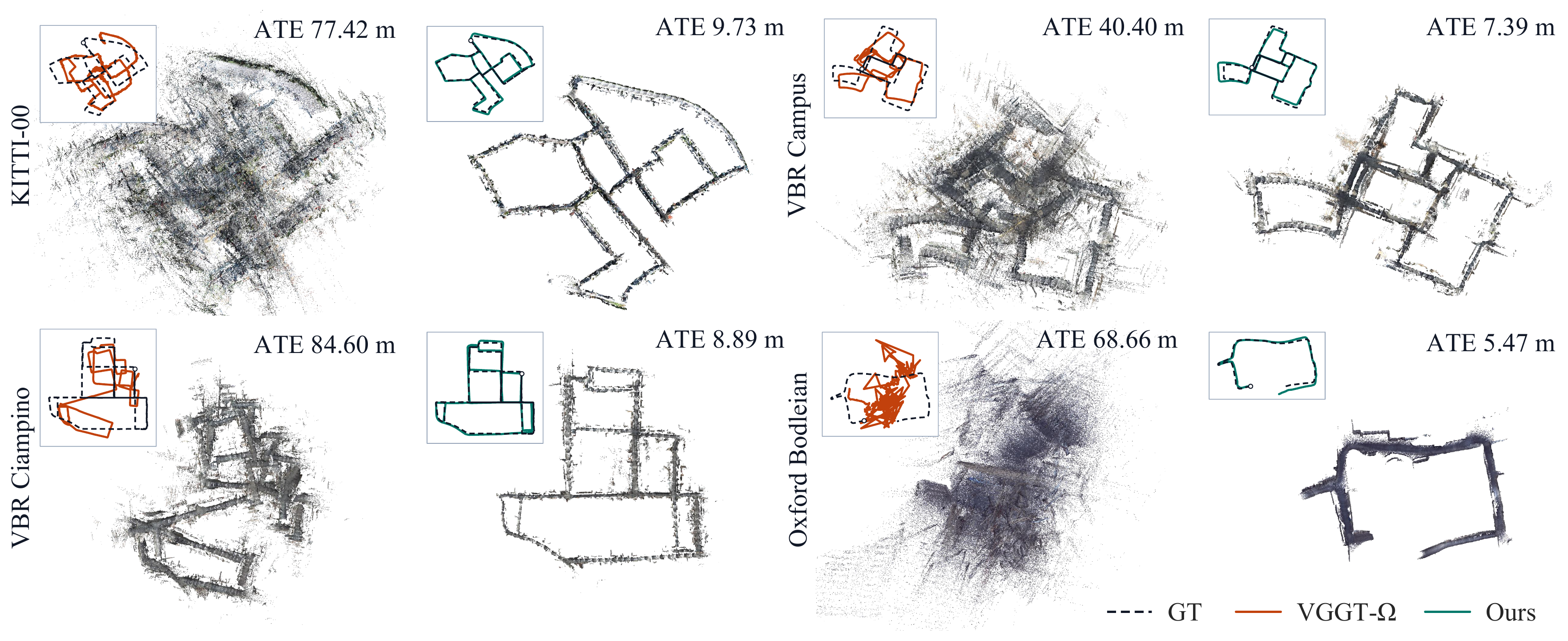}
\vspace{-4mm}
\caption{Trajectory comparison on four long sequences with identical VGGT-Omega points.}
\vspace{-1mm}
\label{fig:traj}
\end{figure*}

Figure~\ref{fig:recon_comparison} compares VGGT-Omega and \method{} (VGGT-Omega) on Oxford {Bodleian Library} and {Keble College}.
Sequential placement drifts across long temporal gaps, reconstructing unclosed loops as offset surfaces several meters apart.
Coarse layout optimization aligns distant revisits and collapses these surfaces onto single facades.
As shown in the local views, shared-camera refinement further reconciles overlapping camera predictions.
Figure~\ref{fig:traj} shows four long sequences with identical VGGT-Omega points.
Native placement stretches streets into duplicated structures, while \method{} recovers closed street networks with lower trajectory error.

\subsection{Ablation Studies}
\label{sec:ablation}

Ablations isolate point sampling, progressive loop scheduling, and placement-motion interaction.

\noindent\textbf{Point sampling bound.}
Table~\ref{tab:layout_design} reports point sampling on cached predictions.
Bounding sampled points cuts graph build and solve time from $12232.18$ to $2004.20$ s ($83.62\%$), while \ate{} changes from $21.7751$ to $21.7569$ m.
Total time includes measurement construction and layout solves.

\begin{table}[ht]
\centering
\caption{Pose AUC and dense geometry on both backends.}
\vspace{1mm}
\label{tab:e2e_omega}
\small
\setlength{\tabcolsep}{3.4pt}
\begin{tabular}{lccccc}
\toprule
\textbf{Method} &
\textbf{AUC@5} $\uparrow$ &
\textbf{AUC@15} $\uparrow$ &
\textbf{AUC@30} $\uparrow$ &
\textbf{Chamfer} $\downarrow$ &
\textbf{F1} $\uparrow$ \\
\midrule
VGGT-Omega & 29.77 & 50.24 & 63.22 & 0.0759 & 65.92 \\
\textbf{\method{}} (VGGT-Omega) & 50.75 & 67.98 & 75.21 & 0.0272 & 91.26 \\
LingBot-MAP & 17.83 & 38.61 & 53.19 & 0.1423 & 61.13 \\
\textbf{\method{}} (LingBot-MAP) & 22.01 & 51.35 & 66.90 & 0.1042 & 74.60 \\
\bottomrule
\end{tabular}
\end{table}

\noindent\textbf{Progressive loop scheduling and solver stabilization.}
Table~\ref{tab:layout_design} compares residual-fit weighting, progressive loop scheduling, and the complete solver.
Progressive loop scheduling reduces graph build and solve time from $2077.49$ to $1477.59$ s.
The complete configuration reaches $17.4708$ m \ate{} in $753.43$ s, with a $41.28$ s final solve.
Progressive scheduling mainly reduces runtime in this comparison, while the complete configuration improves both runtime and trajectory accuracy.
Appendix~\ref{app:layout_controls} details the additional solver components evaluated together in the complete configuration.

\begin{table}[ht]
\centering
\caption{Ablation of point sampling and progressive layout solver on extended sequences.
Rows add components cumulatively.
Appendix~\ref{app:layout_controls} details all test conditions.}
\vspace{1mm}

\label{tab:layout_design}
\small
\begin{tabular}{lrrr}
\toprule
Configuration & \ate{} $\downarrow$ & Graph Build \& Solve (s) $\downarrow$ & Final solve (s) \\
\midrule
Baseline & 21.7751 & 12232.18 & 1404.54 \\
$+$ Bounded sampling & 21.7569 & 2004.20 & 1345.43 \\
$+$ Fit weighting & 21.7770 & 2077.49 & 1387.14 \\
$+$ Progressive loops & 21.5241 & 1477.59 & 688.02 \\
Complete & 17.4708 & 753.43 & 41.28 \\
\bottomrule
\end{tabular}
\end{table}

\begin{table}[ht]
\centering
\caption{Trajectory comparison on identical predictions and one optimized coarse layout.}
\vspace{1mm}

\label{tab:prediction_controls}
\small
\begin{tabular}{l rrr rrr}
\toprule
& \multicolumn{3}{c}{VGGT-Omega} & \multicolumn{3}{c}{LingBot-MAP} \\
\cmidrule(lr){2-4}\cmidrule(lr){5-7}
& \ate{} & \rper{} & \rpet{} & \ate{} & \rper{} & \rpet{} \\
\midrule
Layout Only & 4.9143 & 0.6412 & 0.0817 & 4.0517 & 1.1837 & 0.2126 \\
Single & 4.9079 & 0.6238 & 0.0695 & 4.0256 & 1.1131 & 0.1930 \\
Unary & 4.9067 & 0.6010 & 0.0674 & 4.0192 & 1.0871 & 0.1828 \\
Motion & 11.6776 & \textbf{0.5877} & \textbf{0.0619} & 6.5007 & \textbf{1.0464} & \textbf{0.1816} \\
Shared & \textbf{4.9061} & 0.5899 & 0.0622 & \textbf{4.0168} & 1.0549 & 0.1824 \\
\bottomrule
\end{tabular}
\end{table}

\noindent\textbf{Interaction of placement priors and motion constraints.}
\label{sec:prediction_controls}
Table~\ref{tab:prediction_controls} compares four ways of forming a camera trajectory, Single, Unary, Motion, and Shared, from identical predictions and optimized coarse layouts, with Layout Only as a reference.
Layout Only takes the first available prediction for each frame in chunk order.
Single selects the prediction farthest from its chunk boundaries.
Neither performs frame-level optimization.
Unary uses overlapping pose observations in a robust unary-only objective without relative-motion factors.
Motion uses within-chunk motion factors without placement priors.
Shared combines both factors as in Eq.~\eqref{eq:fine_objective}.
On both backends, Shared reduces relative errors over Single and Unary, with a small \ate{} gain.
Motion yields slightly lower relative errors, but without placement priors, \ate{} increases to $11.678$ and $6.501$ m.
Placement priors preserve the recovered layout, while motion factors improve local smoothness.

\begin{table}[ht]
\centering
\caption{Stage-wise results on identical VGGT-Omega predictions. Layout Only is the direct coarse-layout output; Full adds shared-camera trajectory refinement on the same layout. Oxford and VBR timings are means per sequence. Inference times include all inference required for each output; solver times are reported per stage. Error subscripts give reductions relative to the preceding stage.}
\vspace{1mm}
\label{tab:matched}
\small
\renewcommand{\arraystretch}{1}
\renewcommand{\tabcolsep}{2.5mm}
\resizebox{\linewidth}{!}{%
\begin{tabular}{ll ccc cc}
\toprule
& & \multicolumn{3}{c}{\textbf{Tracking Error}} & \multicolumn{2}{c}{\textbf{Runtime (s)}} \\
\cmidrule(lr){3-5}\cmidrule(lr){6-7}
\textbf{Dataset} & \textbf{Stage} & \textbf{\ate{}} $\downarrow$ & \textbf{\rper{}} ($^\circ$) $\downarrow$ & \textbf{\rpet{}} $\downarrow$ & \textbf{Inference} & \textbf{Stage solve} \\
\midrule
\multirow{3}{*}{\shortstack[l]{\textbf{KITTI}\\($T{=}4541$)}}
& Baseline & $77.42$ & $0.895$ & $2.358$ & $825.745$ & {--} \\
& Layout Only & $9.750_{\scriptscriptstyle(-87\%)}$ & $0.202_{\scriptscriptstyle(-77\%)}$ & $0.131_{\scriptscriptstyle(-94\%)}$ & $896.785$ & $5.40$ \\
& Full & $\mathbf{9.730}_{\scriptscriptstyle(-0.2\%)}$ & $\mathbf{0.154}_{\scriptscriptstyle(-24\%)}$ & $\mathbf{0.093}_{\scriptscriptstyle(-29\%)}$ & $896.785$ & $0.08$ \\
\midrule
\multirow{3}{*}{\shortstack[l]{\textbf{VBR}\\($T{\approx}12{,}383$)}}
& Baseline & $41.91$ & $2.117$ & $0.256$ & $2283.438$ & {--} \\
& Layout Only & $22.44_{\scriptscriptstyle(-46\%)}$ & $0.617_{\scriptscriptstyle(-71\%)}$ & $0.075_{\scriptscriptstyle(-71\%)}$ & $2511.460$ & $200.0$ \\
& Full & $\mathbf{22.44}_{\scriptscriptstyle(-0.0\%)}$ & $\mathbf{0.587}_{\scriptscriptstyle(-5\%)}$ & $\mathbf{0.059}_{\scriptscriptstyle(-22\%)}$ & $2511.460$ & $0.16$ \\
\midrule
\multirow{3}{*}{\shortstack[l]{\textbf{Oxford}\\($T{\approx}320$)}}
& Baseline & $9.861$ & $5.757$ & $4.044$ & $53.297$ & {--} \\
& Layout Only & $1.599_{\scriptscriptstyle(-84\%)}$ & $0.579_{\scriptscriptstyle(-90\%)}$ & $0.173_{\scriptscriptstyle(-96\%)}$ & $54.919$ & $0.03$ \\
& Full & $\mathbf{1.579}_{\scriptscriptstyle(-1\%)}$ & $\mathbf{0.408}_{\scriptscriptstyle(-30\%)}$ & $\mathbf{0.104}_{\scriptscriptstyle(-40\%)}$ & $54.919$ & $0.02$ \\
\bottomrule
\end{tabular}
}
\end{table}

\noindent\textbf{Stage-wise results across layout recovery and trajectory refinement.}
Table~\ref{tab:matched} reports the Layout Only and Full outputs on identical VGGT-Omega predictions.
Layout Only recovers global layout and scale, reducing \ate{} by $87\%$ on KITTI, $84\%$ on Oxford, and $46\%$ on VBR.
Full then adds shared-camera trajectory refinement on that fixed layout.
It reduces \rper{} by $24\%$ on KITTI, $30\%$ on Oxford, and $5\%$ on VBR, and \rpet{} by $29\%$, $40\%$, and $22\%$.

\section{Conclusion}
\label{sec:conclusion}

\method{} turns overlapping local predictions into a globally consistent reconstruction of long video sequences through a hierarchical pose graph.
The method first recovers global layout and scale through a chunk-level $\mathrm{Sim}(3)$ graph, then refines a shared-camera $\mathrm{SE}(3)$ graph with the chunk transforms held fixed.
This separation allows distant loop closures to correct accumulated drift while reconciling conflicting predictions of the same camera across windows.
Evaluations on indoor and outdoor benchmarks show improvements in overall trajectory accuracy and surface reconstruction with both VGGT-Omega and LingBot-MAP.
Extending this hierarchy to online updates and improving its robustness to unreliable loop closures are promising directions for future work.

\bibliography{references}
\bibliographystyle{preprint}

\newpage
\appendix

\section{Appendix}
In this supplementary material, we provide mathematical derivations, implementation details, and extended experimental analyses.
Appendix~\ref{app:jacobian} details closed-form Jacobian derivations and graph coloring.
Appendix~\ref{app:keyframe} specifies the keyframe subgraph and residual interpolation for long sequences.
Appendix~\ref{app:protocol} details public solver settings and optimization hyperparameters.
Appendix~\ref{app:datasets} provides dataset compositions and evaluation splits across all seven benchmarks.
Appendix~\ref{app:long_time} breaks down stage-wise execution times across competing methods.
Appendices~\ref{app:external} and~\ref{app:e2e} report per-dataset tracking and dense 3D geometry results.
Finally, Appendices~\ref{app:windows} through~\ref{app:step_attribution} present window diagnostics, solver ablations, and dense geometry comparisons.

\section{Graph coloring and closed-form Jacobian blocks}
\label{app:jacobian}

This appendix provides mathematical derivations for the sparse Jacobian blocks in Section~\ref{sec:unique}.
Derivative identities follow standard Lie-group calculus~\citep{sola2018micro}.
We first analyze Jacobian sparsity and compare finite-difference graph coloring with analytic derivatives.
We then derive closed-form blocks for $\mathrm{SE}(3)$ residuals and describe automatic differentiation for $\mathrm{Sim}(3)$ residuals.

\noindent\textbf{Jacobian sparsity and manifold structure.}
Nonlinear least-squares iterations linearize $r(x)$ at current poses.
Jacobian $J=\partial r/\partial x$ has one row per residual coordinate and one column per pose parameter.
For a $20{,}000$-frame graph, $n$ reaches $10^5$ and $m$ spans millions.
A dense $J$ consists mostly of zeros.
A relative residual between frames $a$ and $b$ depends only on those two cameras under fixed chunk transforms.

\noindent\textbf{Graph coloring baseline for finite differences.}
Without closed-form derivatives, $J$ can be approximated via finite differences.
Evaluating columns individually costs $n$ residual passes.
Sparsity enables a more efficient approach.
Two columns are structurally orthogonal if they share no residual rows.
Perturbing orthogonal variables simultaneously isolates each column after differencing.
Each residual experiences at most one perturbation~\citep{curtis1974estimation}.
Coloring the column-intersection graph groups mutually orthogonal columns together.
The number of residual sweeps equals the chromatic number.
This depends on graph topology instead of total parameter count.
On a $735$-frame scene, coloring cuts post-processing from $10{,}228$ to $172$ s.
The closed-form derivatives below eliminate these colored sweeps for $\mathrm{SE}(3)$ residuals.

\noindent\textbf{Lie group calculus and rotation identities.}
Let $[u]_\times$ denote the skew-symmetric matrix where $[u]_\times v=u\times v$.
Let $\mathrm{Exp}$ and $\mathrm{Log}$ denote the $\mathrm{SO}(3)$ exponential and logarithm.
We use three standard Lie identities~\citep{sola2018micro}.
\begin{gather}
    \mathrm{Exp}(\phi+\delta)
    \approx
    \mathrm{Exp}(\phi)\,
    \mathrm{Exp}\!\bigl(J_r(\phi)\,\delta\bigr),
    \label{eq:id_jr}
    \\[0.5ex]
    \begin{aligned}
        \mathrm{Log}\!\bigl(R\,\mathrm{Exp}(\delta)\bigr)
        &\approx
        \mathrm{Log}(R)
        + J_r^{-1}\!\bigl(\mathrm{Log}(R)\bigr)\,\delta,
        \\
        \mathrm{Log}\!\bigl(\mathrm{Exp}(\delta)\,R\bigr)
        &\approx
        \mathrm{Log}(R)
        + J_l^{-1}\!\bigl(\mathrm{Log}(R)\bigr)\,\delta,
    \end{aligned}
    \label{eq:id_log}
    \\[0.5ex]
    A\,\mathrm{Exp}(u)
    =
    \mathrm{Exp}(Au)\,A,
    \qquad
    A\in\mathrm{SO}(3).
    \label{eq:id_adj}
\end{gather}
The right Jacobian and its inverse have the closed forms
($\theta=\lVert\phi\rVert$)
\begin{equation}
    J_r(\phi)=I-\tfrac{1-\cos\theta}{\theta^2}[\phi]_\times
    +\tfrac{\theta-\sin\theta}{\theta^3}[\phi]_\times^2,
    \qquad
    J_r^{-1}(\phi)=I+\tfrac12[\phi]_\times
    +\Big(\tfrac1{\theta^2}-\tfrac{1+\cos\theta}{2\theta\sin\theta}\Big)
    [\phi]_\times^2,
\end{equation}
and $J_l^{-1}(\omega)=J_r^{-1}(\omega)^{\!\top}$.
Identity~\eqref{eq:id_jr} differentiates global rotation vectors, where frame $t$ carries $x_t=(\phi_t,\tau_t)\in\mathbb{R}^6$ with $R_t=\mathrm{Exp}(\phi_t)$.

\noindent\textbf{Closed-form factorized $\mathrm{SE}(3)$ residual.}
Consider a fine edge between source $a$ and target $b$ with measurement $M=(R_M,t_M)$.
Expanding $M^{-1}F_b^{-1}F_a$ into rotation and translation  yields
\begin{equation}
    r_R=\mathrm{Log}\!\big(\underbrace{R_M^{\!\top}R_b^{\!\top}R_a}_{E_R}\big),
    \qquad
    r_t=R_M^{\!\top}\!\big(R_b^{\!\top}(\tau_a-\tau_b)-t_M\big).
    \label{eq:split_res}
\end{equation}
This factorization provides favorable sparsity.
Specifically, $r_R$ never depends on translation, while $r_t$ never depends on source rotation.
A dense $6{\times}12$ local Jacobian exceeds residual requirements.
After the coarse stage, $M$ remains constant.

\noindent\textbf{Analytic rotation derivatives.}
Perturbing $\phi_a\to\phi_a+\delta$ and applying \eqref{eq:id_jr} gives
$E_R\to E_R\,\mathrm{Exp}(J_r(\phi_a)\delta)$, so by \eqref{eq:id_log}
\begin{equation}
    \frac{\partial r_R}{\partial\phi_a}
    = J_r^{-1}(r_R)\,J_r(\phi_a).
\end{equation}
Perturbing $\phi_b$ gives
$R_b^{\!\top}\to\mathrm{Exp}(-J_r(\phi_b)\delta)R_b^{\!\top}$. Transporting
the perturbation through $R_M^{\!\top}$ with \eqref{eq:id_adj} yields
$E_R\to\mathrm{Exp}(-R_M^{\!\top}J_r(\phi_b)\delta)\,E_R$ and hence
\begin{equation}
    \frac{\partial r_R}{\partial\phi_b}
    = -\,J_l^{-1}(r_R)\,R_M^{\!\top}J_r(\phi_b).
\end{equation}
Each block multiplies at most three $3{\times}3$ matrices.
These include the inverse log Jacobian, a frame-transport rotation, and the target right Jacobian.

\noindent\textbf{Analytic translation derivatives.}
The translation residual is linear in translations.
\begin{equation}
    \frac{\partial r_t}{\partial\tau_a}=R_M^{\!\top}R_b^{\!\top},
    \qquad
    \frac{\partial r_t}{\partial\tau_b}=-R_M^{\!\top}R_b^{\!\top},
\end{equation}
depending on rotation only through the target.
Using $-[u]_\times v=[v]_\times u$,
\begin{equation}
    \frac{\partial r_t}{\partial\phi_b}
    = R_M^{\!\top}\big[R_b^{\!\top}(\tau_a-\tau_b)\big]_\times J_r(\phi_b),
    \qquad
    \frac{\partial r_t}{\partial\phi_a}=0.
\end{equation}
A fine edge contributes five nonzero $3{\times}3$ blocks.
The prior residual from Eq.~\eqref{eq:prior} with placement $Q=(R_Q,t_Q)$ factorizes similarly.
Specifically, $r^p_R=\mathrm{Log}(R_Q^{\!\top}R_g)$ and $r^p_t=R_Q^{\!\top}(\tau_g-t_Q)$.
This yields two diagonal blocks.
\begin{equation}
    \frac{\partial r^p_R}{\partial\phi_g}=J_r^{-1}(r^p_R)\,J_r(\phi_g),
    \qquad
    \frac{\partial r^p_t}{\partial\tau_g}=R_Q^{\!\top}.
\end{equation}

\noindent\textbf{Sparsity pattern assembly and reuse.}
Quantities $R_t$ and $J_r(\phi_t)$ appear in every factor incident to frame $t$.
They are computed once per iteration in batch and shared.
Per-edge terms $J_r^{-1}(r_R)$ and $J_l^{-1}(r_R)$ are evaluated in batch from current residuals.
The CSR sparsity structure depends only on graph topology, requiring a single initialization.
Subsequent iterations update only numeric block values.
Jacobian construction requires only a few batched $3{\times}3$ multiplications per edge, scaling linearly without residual sweeps.

\noindent\textbf{Coarse $\mathrm{Sim}(3)$ residual formulation.}
Let $A_i=(s_i,R_i,t_i)$ and $A_j=(s_j,R_j,t_j)$.
Relative similarity $\widehat{Z}_{ij}=(\hat s,\hat R,\hat t)$ comes from confidence-weighted Umeyama alignment on overlap or loop pointmaps.
The coarse residual in Eq.~\eqref{eq:coarse} is
\begin{equation}
    E_{ij}=\widehat{Z}_{ij}^{-1}A_j^{-1}A_i,\qquad
    \widetilde{r}^{c}_{ij}=\sqrt{w_{ij}}\,r^{c}_{ij}=\sqrt{w_{ij}}\,
    \begin{bmatrix}
    \mathrm{Log}_{\mathrm{SO}(3)} R(E_{ij})\\
    \tau(E_{ij})\\
    \log s(E_{ij})
    \end{bmatrix}\in\mathbb{R}^{7},
\end{equation}
ordered as $(\omega,\,\tau,\,\log s)$ with $\omega\in\mathbb{R}^3$ and translation $\tau\in\mathbb{R}^3$.
Edge weights apply prior to the Huber penalty.
Fixing $A_1$ leaves $7(K-1)$ active parameters.
Huber loss is coordinate-wise with $\delta=1$.
Adaptive Jacobi scaling normalizes parameter updates across rotation, translation, and scale.

\noindent\textbf{Coarse $\mathrm{Sim}(3)$ blocks via automatic differentiation.}
Similarity composition couples scales, rotations, and translations between endpoints.
We differentiate per-edge residuals using vectorized float64 automatic differentiation.
This outputs one $7{\times}14$ block per edge, assembled into CSR format.
The closed-form $\mathrm{SE}(3)$ blocks and automatically differentiated $\mathrm{Sim}(3)$ blocks match numerical derivatives within $1.1\times 10^{-8}$.
Because the fine stage holds $A^{\star}$ fixed, chunk columns drop before the $\mathrm{SE}(3)$ solve.

\section{Long-sequence keyframe subgraph}
\label{app:keyframe}

For sequences exceeding $2000$ frames, optimizing camera variables at every single frame is computationally unnecessary.
Camera motion across adjacent video frames is temporally smooth.
Therefore, the fine stage solves the objective over an active keyframe subgraph.
Dense per-frame corrections are recovered by interpolating their rotation and translation components.

\noindent\textbf{Keyframe selection rules.}
Let $T$ denote frame count and $s{=}5$ the default stride. Keyframe selection in this appendix uses zero-based frame indices.
Target indices include $\{0,s,2s,\ldots\}$ and $\{T{-}1\}$.
Each chunk $[a,b)$ also contributes boundary frames $a$ and $b{-}1$.
For adjacent chunk overlap $\mathcal{O}=\{t:t\in C_k\cap C_{k+1}\}$, we check existing samples.
If fewer than $m=\min(3,\lvert\mathcal{O}\rvert)$ indices lie in $\mathcal{O}$, we insert $m$ linearly spaced frames.
Within-chunk edges connect consecutive keyframes, and placement priors attach only to retained nodes.

\noindent\textbf{Rotation and translation interpolation of pose corrections.}
Let $Q_t$ denote the first-occurrence coarse pose for frame $t$ in chunk order.
A solved keyframe pose $F_{t_i}$ defines correction $\Delta_{t_i}=F_{t_i}Q_{t_i}^{-1}$ in world coordinates.
Between keyframes $t_i<t_{i+1}$ with weight $\alpha=(t-t_i)/(t_{i+1}-t_i)$,
\begin{equation}
  R(\Delta_t)=\mathrm{Slerp}\big(R(\Delta_{t_i}),R(\Delta_{t_{i+1}}),\alpha\big),
  \qquad
  \tau(\Delta_t)=(1-\alpha)\,\tau(\Delta_{t_i})+\alpha\,\tau(\Delta_{t_{i+1}}),
  \label{eq:interp}
\end{equation}
and $F_t=\Delta_t\,Q_t$.

\noindent\textbf{Stride sensitivity and comparisons.}
Overlap averaging computes consensus poses before applying residual interpolation, while omitting the fine optimization stage.
Table~\ref{tab:stride} compares uniform $s{=}10$ with motion-spaced $s{=}10$.
Motion spacing allocates targets by accumulated motion instead of frame indices.
Table~\ref{tab:stride_long} further extends this evaluation across three long sequences.
Table~\ref{tab:dense_kf} evaluates dense versus keyframe graphs on the KITTI dataset.
\begin{table}[ht]
\centering
\caption{Keyframe subgraph on KITTI after the same coarse $A^{\star}$.}
\vspace{1mm}
\label{tab:stride}
\small
\setlength{\tabcolsep}{3.4pt}
\begin{tabular}{lccccc}
\toprule
\textbf{Refinement graph} &
\ate{} $\downarrow$ &
\rper{} $\downarrow$ &
\rpet{} $\downarrow$ &
$|\mathcal{K}|$ &
$t_{\mathrm{f}}$ (s) \\
\midrule
Overlap averaging & $9.735$ & $0.202$ & $0.131$ & --- & --- \\
Uniform stride $10$ & $9.728$ & \stgv{0.158}{-22\%} & \stgv{0.094}{-28\%} & $605$ & $0.055$ \\
Motion stride $10$ & $9.729$ & \stgv{0.156}{-23\%} & \stgv{0.094}{-28\%} & $725$ & $0.104$ \\
Uniform stride $5$ & $9.730$ & \stgv{0.154}{-24\%} & \stgv{0.093}{-29\%} & $1059$ & $0.075$ \\
\bottomrule
\end{tabular}
\vspace{-1mm}
\end{table}

\begin{table}[ht]
\centering
\caption{The same keyframe comparison on additional long sequences.}
\vspace{1mm}
\label{tab:stride_long}
\small
\setlength{\tabcolsep}{3.2pt}
\begin{tabular}{lrcccccc}
\toprule
& &
\multicolumn{2}{c}{Consensus} &
\multicolumn{2}{c}{Uniform $s{=}10$} &
\multicolumn{2}{c}{Uniform $s{=}5$} \\
\cmidrule(lr){3-4}\cmidrule(lr){5-6}\cmidrule(lr){7-8}
\textbf{Scene} & $T$ &
\rper{} & \rpet{} &
\rper{} & \rpet{} &
\rper{} & \rpet{} \\
\midrule
campus\_train0 & $12{,}042$
& $0.204$ & $0.129$
& $0.122$ & $0.118$
& $0.119$ & $0.117$ \\
spagna\_train0 & $14{,}141$
& $0.542$ & $0.048$
& $0.535$ & $0.035$
& $0.535$ & $0.034$ \\
downtown2 & $2200$
& $3.509$ & $0.083$
& $3.507$ & $0.082$
& $3.507$ & $0.082$ \\
\bottomrule
\end{tabular}
\vspace{-1mm}
\end{table}

Uniform stride $s{=}5$ yields comparable or lower relative errors than $s{=}10$ in these comparisons.
On KITTI, solve times are $0.075$ and $0.055$ s, respectively.

\begin{table}[ht]
\centering
\caption{Dense fine graph versus uniform stride $5$ on VGGT-Omega predictions and coarse $A^{\star}$.}
\vspace{1mm}
\label{tab:dense_kf}
\small
\setlength{\tabcolsep}{3.2pt}
\begin{tabular}{lrcccccc}
\toprule
& &
\multicolumn{3}{c}{Dense $F_t$} &
\multicolumn{3}{c}{Uniform $s{=}5$} \\
\cmidrule(lr){3-5}\cmidrule(lr){6-8}
\textbf{Scene} & $T$ &
\ate{} & \rper{} & \rpet{} &
\ate{} & \rper{} & \rpet{} \\
\midrule
Courthouse & $1106$
& $2.599$ & $10.82$ & $0.517$
& $2.600$ & $10.81$ & $0.517$ \\
downtown1 & $1427$
& $3.223$ & $3.940$ & $0.0875$
& $3.223$ & $3.940$ & $0.0876$ \\
KITTI & $4541$
& $9.731$ & $0.144$ & $0.086$
& $9.730$ & $0.154$ & $0.093$ \\
\bottomrule
\end{tabular}
\vspace{-1mm}
\end{table}

Table~\ref{tab:dense_kf} compares dense and keyframe graphs.
On Courthouse and downtown1 ($1106$ and $1427$ frames), stride $5$ yields errors comparable to the dense graph.
On KITTI ($4541$ frames), dense solving improves \rper{} and \rpet{} by $6\%$ and $8\%$ with unchanged \ate{}.
The default configuration retains the keyframe subgraph on long sequences.

\section{Hyperparameters and solver settings}
\label{app:protocol}

Table~\ref{tab:hparams} lists the default settings for the main benchmark.

\begin{table}[t]
\centering
\caption{Solver hyperparameters and optimization settings across backends.
We use the public configurations of VGGT-Omega~\citep{wang2026vggtomega},
LingBot-MAP~\citep{chen2026lingbotmap}, and
SALAD~\citep{izquierdo2024salad}.
}
\vspace{1mm}
\renewcommand{\arraystretch}{1.3}
\label{tab:hparams}
\small
\begin{tabularx}{\linewidth}{@{}
  >{\raggedright\arraybackslash}p{0.3\linewidth}
  >{\raggedright\arraybackslash}X
  @{}}
\toprule
\textbf{Quantity} & \textbf{Value} \\
\midrule
VGGT-Omega default window
  & $C{=}60$, $O{=}30$ frames \\
LingBot-MAP default window
  & $C{=}120$, $O{=}32$ frames, windowed inference \\
Loop window
  & $20$ frames per endpoint \\
SALAD retrieval
  & similarity $0.85$, top-$5$, NMS $25$ frames, minimum chunk separation $1$, one pair per chunk pair (first hit) \\
Loop-constraint addition
  & $4$ configured span-ordered stages, expanded to at most $8$ by adaptive continuation \\
Keyframe subgraph ($T>2000$)
  & uniform stride $5$, all chunk endpoints, at least $\min(3,|\mathcal O|)$ frames in each overlap $\mathcal O$, correction interpolation (Appendix~\ref{app:keyframe}) \\
Refinement weights (Eq.~\eqref{eq:fine_objective})
  & $\lambda_f{=}1$, $\lambda_p{=}0.25$, fixed across backends and windows \\
Coarse $\mathrm{Sim}(3)$
  & Huber $f_{\mathrm{scale}}{=}1$, adaptive Jacobi scaling (\texttt{x\_scale}${=}$`\texttt{jac}'), \texttt{max\_nfev}${=}5000$, \texttt{ftol}${=}$\texttt{xtol}${=}$\texttt{gtol}${=}10^{-8}$ (SciPy defaults) \\
Coarse continuation
  & intermediate stages ($100$ evals), adaptive full solve ($500$), full fallback ($5000$) \\
Active-bound restarts
  & at most $2$ passes of $1000$ evaluations, initial log-scale bounds $\pm\log 100$, recentered on restart \\
Shared-camera $\mathrm{SE}(3)$
  & fixed $A^\star$, Huber scale $\delta=1$, \texttt{max\_nfev}${=}100$, \texttt{ftol}${=}$\texttt{xtol}${=}$\texttt{gtol}${=}10^{-8}$, \texttt{translation\_scale}${=}1$ \\
Jacobian
  & closed-form $3{\times}3$ $\mathrm{SE}(3)$ blocks in CSR, coarse $\mathrm{Sim}(3)$ by float64 automatic differentiation \\
\bottomrule
\end{tabularx}
\end{table}

\noindent\textbf{Backend inputs.}
The graph uses camera poses and pointmaps from each inference chunk.
When a backend provides depth rather than pointmaps, we unproject its depth predictions into 3D points.
For LingBot-MAP, we decode camera-to-world poses and unproject depth within inference windows.

\noindent\textbf{Trajectory evaluation.}
We compute trajectory errors with evo, enabling \texttt{correct\_scale} for the per-sequence $\mathrm{Sim}(3)$ alignment used to evaluate \ate{}.

\noindent\textbf{Point sampling and pairwise alignment.}
For adjacent alignment, we sample at most $2\times10^5$ point pairs evenly from the flattened arrays, excluding invalid points and pairs with non-positive confidence scores.
Paired confidence scores are combined by their geometric mean.
Alignment uses at most $10$ IRLS iterations with an adaptive Huber threshold.

\noindent\textbf{Loop-constraint weighting.}
The alignment fit statistic $q_{ij}$ is the larger of the two endpoint IRLS residual $95$th percentiles ($p_{95}$).
If $p_{95}$ is unavailable at either endpoint, we use the larger endpoint $90$th percentile ($p_{90}$).
We convert this statistic to a bounded edge weight,
\[
    w_{ij}=\operatorname{clip}\!\left(
        \frac{\sigma}{\sigma+\max(q_{ij},0)},\,w_{\min},\,w_{\max}
    \right),
\]
where $\operatorname{clip}(x,a,b)=\min\{b,\max\{a,x\}\}$ restricts the weight to the interval $[a,b]$.
We use $\sigma=0.05$, $w_{\min}=0.05$, and $w_{\max}=1$.

\noindent\textbf{Camera initialization and placement priors.}
For each frame $t$, we select its first occurrence in chunk order and apply the coarse placement to obtain $Q_t$.
We initialize the shared camera as $F_t=Q_t$ and use $Q_t$ as the reference for correction interpolation in Appendix~\ref{app:keyframe}.
For every frame retained in the optimization, each chunk prediction $Q_{k,i}$ with $g(k,i)=t$ contributes a separate placement prior to the same shared camera $F_t$, as defined in Eq.~\eqref{eq:prior}.
The first camera is fixed to its coarse placement as the coordinate reference.

\noindent\textbf{Solver configuration.}
Each free chunk is parameterized by an absolute rotation vector, a translation vector, and log-scale.
Coarse graph optimization uses bounded trust-region reflective least squares.
Adaptive Jacobi scaling accounts for differences in parameter sensitivity across rotation, translation, and scale.
Initial log-scale bounds are centered on the sequential initialization.
The fine stage holds the recovered chunk transformations fixed and optimizes camera poses, with the first camera held as the reference.
Table~\ref{tab:hparams} gives the loss scales, residual weights, evaluation budgets, and termination tolerances.
Keyframe selection and correction interpolation follow Appendix~\ref{app:keyframe}.

\noindent\textbf{Adaptive schedule and restarts.}
Let $L$ denote the number of loop edges, $n_{\min}$ the shortest span count, and $S=4$ the configured stage count.
When loop spans vary and $n_{\min}S\geq L(S-1)$, the schedule expands to $\min(2S,L)$ stages.
Longer-span constraints then enter through additional intermediate layouts.
Intermediate stages and the adaptive full-graph solve use the separate budgets in Table~\ref{tab:hparams}.
If the full solve reaches its evaluation ceiling without converging, the solver invokes the full-budget fallback.
Active scale bounds trigger the restart routine instead.
Restarts warm-start from the latest estimate and recenter scale bounds.
They continue while the problem remains unconverged and bounds remain active, subject to the restart limits in Table~\ref{tab:hparams}.

\section{Datasets and splits}
\label{app:datasets}

We evaluate \method{} across seven public benchmarks spanning $61$ diverse sequences.
Table~\ref{tab:eval_data} lists dataset compositions, sensor domains, and evaluation properties.

\begin{table}[t]
\centering
\caption{Dataset composition and experimental splits.}
\vspace{1mm}
\renewcommand{\arraystretch}{1.15}
\label{tab:eval_data}
\small
\setlength{\tabcolsep}{3.4pt}
\begin{tabular}{llrl}
\toprule
\textbf{Dataset} & \textbf{Domain} & \textbf{\#Seq.} & \textbf{What it tests} \\
\midrule
7-Scenes & Indoor RGB-D
  & \multirow{3}{*}{38} & Textureless rooms, motion blur \\
ETH3D & Indoor and outdoor
  & & Mixed-scale laser-scanned scenes \\
Neural RGB-D & Indoor RGB-D
  & & Cluttered rooms, fine geometry \\
Tanks and Temples & Outdoor structures
  & 6 & Large objects and buildings \\
Oxford Spires & Campus indoor and outdoor
  & 9 & Loops and scale change ($T{\approx}320$) \\
KITTI seq00 & Driving
  & 1 & Long outdoor loop ($T{=}4541$) \\
VBR & City-scale Rome
  & 7 & Extreme length ($T$ up to $18{,}846$) \\
\midrule
Total & & 61 & \\
\bottomrule
\end{tabular}
\end{table}

\noindent\textbf{Indoor room-scale and multi-scale sequences.}
These three datasets evaluate performance across local and multi-scale settings with varying sequence lengths and scene complexity.
With the default VGGT-Omega chunk size of $60$ frames and overlap size of $30$ frames, $10$ of the $11$ ETH3D still sequences fit within a single chunk.
For these sequences, the layout and shared-camera graphs remain inactive, and the Native, Layout Only, and Full trajectories coincide.
Neural RGB-D spans $2$--$11$ chunks per sequence, while 7-Scenes spans $3$--$6$.
\textbf{7-Scenes}~\citep{shotton2013scene} consists of seven Kinect-captured rooms: Chess, Fire, Heads, Office, Pumpkin, Kitchen, and Stairs.
\textbf{ETH3D}~\citep{schops2017eth3d} includes high-resolution indoor and outdoor still sequences with laser-scanner ground truth, covering offices, reliefs, facades, playgrounds, and gardens.
\textbf{Neural RGB-D}~\citep{azinovic2022neural} contains nine room-scale indoor scenes with structured-light depth, including kitchens, staircases, and apartments.

\noindent\textbf{Outdoor structures and buildings.}
\textbf{Tanks and Temples}~\citep{knapitsch2017tanks} evaluates Barn, Caterpillar,
Courthouse, Ignatius, Meeting Room, and Truck.

\noindent\textbf{Extended traversals and driving loops.}
\textbf{Oxford Spires}~\citep{tao2024spires} provides nine campus sequences with
LiDAR-inertial ground truth, subsampled to $T{\approx}320$ frames.
The sequences mix outdoor courtyards, interiors, and staircases, with
revisits after long temporal gaps.
\textbf{KITTI}~\citep{geiger2012kitti} uses odometry sequence~00 only
($T{=}4541$), a standard benchmark for long driving loops.
\textbf{VBR}~\citep{brizi2024vbr} contains seven long urban sequences through Rome,
$8{,}815$--$18{,}846$ frames per sequence (dataset-mean $T{=}12{,}383$).

Additional keyframe and layout-solver experiments use the downtown1, downtown2, and downtown5 sequences from Droid-W~\citep{li2026droidw}.

\noindent\textbf{Aggregation.}
Trajectory benchmark metrics are averaged within each dataset and then equally across the seven datasets.
Dense geometry metrics use an equal-weight average over the $38$ evaluated sequences, as detailed in Appendix~\ref{app:e2e}.

\section{Reported pipeline-stage timings}
\label{app:long_time}

Table~\ref{tab:long_time} reports pipeline component times with inference on a single GPU and \method{} graph solves on CPU.
The \method{} timings are calculated from Table~\ref{tab:matched}, with $t_{\mathrm{loop}}=t_{\mathrm{g}}-t_{\mathrm{g}}^{\mathrm{N}}$ and $t_{\mathrm{solve}}=t_{\mathrm{c}}+t_{\mathrm{f}}$.
For VGGT-Long, reported components include chunk inference, loop-window inference when enabled, and pairwise $\mathrm{Sim}(3)$ alignment.
We report VGGT-Long with and without loops on the default VGGT-Omega windows, alongside its released configuration.
GlueMap component times include Pi3 star forwards, tracking, and coarse mapping on one GPU.

\begin{table}[t]
\centering
\caption{Mean pipeline component times (s) on Oxford, KITTI, and VBR. For \method{}, component sums include chunk and loop inference and coarse and fine solves.}
\vspace{1mm}
\label{tab:long_time}
\small
\setlength{\tabcolsep}{6.0pt}
\resizebox{\linewidth}{!}{%
\begin{tabular}{ll rrr}
\toprule
& & \multicolumn{3}{c}{\textbf{Split}} \\
\cmidrule(lr){3-5}
\textbf{Method} & \textbf{Component} & \textbf{Oxford} & \textbf{KITTI seq00} & \textbf{VBR} \\
& & ($T{\approx}320$) & ($T{=}4541$) & ($T{\approx}12{,}383$) \\
\midrule
\multirow{4}{*}{\shortstack[l]{\textbf{\method{}}\\(Ours)}}
& Local baseline chunk inference ($t_{\mathrm{g}}^{\mathrm{N}}$) & $53.3$ & $825.7$ & $2283.4$ \\
& Loop-window inference ($t_{\mathrm{loop}}$) & $1.6$ & $71.0$ & $228.0$ \\
& CPU coarse + fine solves ($t_{\mathrm{solve}}$) & $0.05$ & $5.5$ & $200.2$ \\
\cmidrule(lr){2-5}
& Component sum & $55.0$ & $902.3$ & $2711.6$ \\
\midrule
\multirow{4}{*}{\shortstack[l]{\textbf{VGGT-Long}\\\textit{(w/ loops), on}\\\textit{Omega windows}}}
& Chunk inference & $57.2$ & $924.2$ & $2664.0$ \\
& Loop-window inference & $0.5$ & $75.3$ & $252.7$ \\
& Chunk alignment ($\mathrm{Sim}(3)$ IRLS) & $139.9$ & $2882.0$ & $9489.5$ \\
\cmidrule(lr){2-5}
& Component sum & $197.6$ & $3881.5$ & $12{,}406.2$ \\
\midrule
\multirow{3}{*}{\shortstack[l]{\textbf{VGGT-Long}\\\textit{(w/o loops), on}\\\textit{Omega windows}}}
& Chunk inference & $67.5$ & $910.1$ & $2919.0$ \\
& Chunk alignment ($\mathrm{Sim}(3)$ IRLS) & $185.6$ & $2457.1$ & $8237.7$ \\
\cmidrule(lr){2-5}
& Component sum & $253.1$ & $3367.2$ & $11{,}156.7$ \\
\midrule
\multirow{3}{*}{\shortstack[l]{\textbf{VGGT-Long}\\\textit{(released)}}}
& Chunk inference & $67.4$ & $794.6$ & $2329.4$ \\
& Chunk alignment ($\mathrm{Sim}(3)$ IRLS) & $169.2$ & $2228.8$ & $5036.7$ \\
\cmidrule(lr){2-5}
& Component sum & $236.6$ & $3023.3$ & $7366.2$ \\
\midrule
\multirow{4}{*}{\shortstack[l]{\textbf{GlueMap}\\\textit{(Pi3 coarse)}}}
& Star forward (Pi3) & $280.9$ & $2913.2$ & $7803.2$ \\
& VGGSfM tracking & $996.2$ & $11{,}947.8$ & $38{,}762.2$ \\
& Coarse mapping & $215.2$ & $1935.8$ & $5077.2$ \\
\cmidrule(lr){2-5}
& Component sum & $1492.3$ & $16{,}796.8$ & $51{,}642.6$ \\
\bottomrule
\end{tabular}%
}
\end{table}

The large alignment times in VGGT-Long show pairwise 3D point alignment to be a dominant computational bottleneck.
Contrastly, \method{} accelerates pairwise alignment via bounded sampling while using a fast sparse graph solver.
Dense point tracking accounts for most of GlueMap runtime.

\section{Per-dataset comparison}
\label{app:external}

Table~\ref{tab:external} in the main text summarizes macro-averaged tracking metrics across all seven public benchmarks.
Tables~\ref{tab:app_ate}, \ref{tab:app_rperot}, and~\ref{tab:app_rpetrans} report detailed per-dataset tracking errors for \ate{}, \rper{}, and \rpet{}, respectively.
In each column, bold and underline mark the best and second-best results across all evaluated methods.
On VGGT-Omega, \method{} improves relative rotation and translation accuracy across all seven datasets.
Trajectory \ate{} improvements are largest on extended outdoor traversals (KITTI and VBR).
In these environments, global loop constraints effectively eliminate accumulated drift.
On compact indoor scenes, baseline predictions already achieve centimeter-level trajectory accuracy.
For LingBot-MAP, \method{} acts as a modular post-processing method.
It reduces macro-averaged \ate{} from $8.288$ to $4.017$ m.

\begin{table}[ht]
\centering
\caption{Per-dataset \ate{} on the seven public datasets
(lower is better). NRGBD is Neural RGB-D, 7S is 7-Scenes,
T\&T is Tanks and Temples. Bold and underline mark the full-table best and second-best values in each column.}
\vspace{1mm}
\label{tab:app_ate}
\scriptsize
\setlength{\tabcolsep}{3.2pt}
\resizebox{\linewidth}{!}{%
\begin{tabular}{lrrrrrrr}
\toprule
\textbf{Method} & \textbf{7S} & \textbf{ETH} &
\textbf{KITTI} & \textbf{NRGBD} & \textbf{Oxford} &
\textbf{T\&T} & \textbf{VBR} \\
\midrule
VGGT-Omega~\citep{wang2026vggtomega} & \textbf{0.0370} & 0.118 &
77.42 & \textbf{0.0190} & 9.861 & 0.518 & 41.914 \\
$\pi^3$~\citep{wang2025pi3} & 0.0574 & \underline{0.104} &
68.22 & 0.0244 & 12.06 & \textbf{0.467} & 87.269 \\
GlueMap~\citep{pan2026gluemap} & 0.0624 & 0.810 &
193.5 & 0.0232 & 16.65 & 0.521 & 105.4 \\
Scal3R~\citep{xie2026scal3r} & 0.0720 & 0.236 & 86.25 & 0.0471 & 4.573 &
0.572 & 101.535 \\
VGGT-Long~\citep{deng2025vggtlong} & 0.0743 & 0.204 & 48.84 & 0.0374 & 9.451 &
0.550 & 99.832 \\
VGGT-Align~\citep{zhang2026vggtalign} & 0.0731 & 0.209 & 69.80 & 0.0374 & 8.567 &
0.537 & 99.867 \\
LingBot-MAP~\citep{chen2026lingbotmap} & 0.0789 & 0.441 & 23.40 &
0.0554 & 2.391 & 0.518 &
31.129 \\
ZipMap~\citep{jin2026zipmap} & 0.0702 & 0.236 & 127.22 & 0.0735 & 28.33 &
0.692 & 40.987 \\
VGGT~\citep{wang2025vggt} & 0.0705 & 0.237 & 79.87 & 0.164 & 21.40 &
0.536 & 44.123 \\
LoGeR~\citep{zhang2026loger} & 0.118 & 0.370 & 174.96 & 0.138 & 5.522 &
1.078 & 102.650 \\
TTT3R~\citep{chen2026ttt3r} & 0.104 & 1.189 & 46.93 & 0.191 & 12.51 &
1.302 & 104.199 \\
CUT3R~\citep{wang2025cut3r} & 0.296 & 1.376 & 190.19 & 0.546 & 16.26 &
1.839 & 103.748 \\
\midrule
\textbf{\method{} (VGGT-Omega)} & \underline{0.0381} & \textbf{0.0546} &
\underline{9.730} & \underline{0.0219} & \textbf{1.579} &
\underline{0.476} & \underline{22.443} \\
\textbf{\method{} (LingBot-MAP)} & 0.0901 & 0.488 & \textbf{6.063} &
0.126 & \underline{2.346} & 1.026 & \textbf{17.978} \\
\bottomrule
\end{tabular}}
\end{table}

\begin{table}[ht]
\centering
\caption{Per-dataset \rper{} on the same seven datasets (lower is
better). Bold and underline mark the full-table best and
second-best values in each column.}
\vspace{1mm}
\label{tab:app_rperot}
\scriptsize
\setlength{\tabcolsep}{3.2pt}
\resizebox{\linewidth}{!}{%
\begin{tabular}{lrrrrrrr}
\toprule
\textbf{Method} & \textbf{7S} & \textbf{ETH} &
\textbf{KITTI} & \textbf{NRGBD} & \textbf{Oxford} &
\textbf{T\&T} & \textbf{VBR} \\
\midrule
VGGT-Omega~\citep{wang2026vggtomega} & \underline{0.501} & 0.868 & 0.895 &
0.146 & 5.757 & 2.458 & 2.117 \\
$\pi^3$~\citep{wang2025pi3} & 0.510 & \underline{0.729} & 0.218 &
0.0999 & 6.912 & 2.036 & \underline{0.636} \\
GlueMap~\citep{pan2026gluemap} & 0.507 & 1.755 & 0.535 &
0.0984 & 4.337 & 2.308 & 1.388 \\
Scal3R~\citep{xie2026scal3r} & 0.520 & 2.036 & \underline{0.215} &
\underline{0.0957} & 2.193 & \underline{1.985} & 2.589 \\
VGGT-Long~\citep{deng2025vggtlong} & 0.535 & 1.257 & 0.264 &
0.118 & 5.135 & 2.027 & 0.806 \\
VGGT-Align~\citep{zhang2026vggtalign} & 0.536 & 1.258 & 2.141 &
0.118 & 5.109 & 2.027 & 0.808 \\
LingBot-MAP~\citep{chen2026lingbotmap} & 0.579 & 3.345 & 1.051 & 0.254 & 1.106 &
2.162 & 4.226 \\
ZipMap~\citep{jin2026zipmap} & 0.632 & 1.098 & 1.495 & 0.161 & 11.84 &
2.531 & 1.786 \\
VGGT~\citep{wang2025vggt} & 0.605 & 1.284 & 2.186 & 0.232 & 12.30 &
2.849 & 2.656 \\
LoGeR~\citep{zhang2026loger} & 0.702 & 1.791 & 2.552 & 0.407 & 2.185 &
3.043 & 3.919 \\
TTT3R~\citep{chen2026ttt3r} & 0.711 & 21.19 & 0.441 & 0.305 & 4.798 &
3.457 & 4.638 \\
CUT3R~\citep{wang2025cut3r} & 0.907 & 10.91 & 6.485 & 0.851 & 4.644 &
2.979 & 4.510 \\
\midrule
\textbf{\method{} (VGGT-Omega)} & \textbf{0.468} & \textbf{0.476} &
\textbf{0.154} & \textbf{0.0710} & \textbf{0.408} &
\textbf{1.965} & \textbf{0.587} \\
\textbf{\method{} (LingBot-MAP)} & 0.586 & 2.287 & 0.275 &
0.280 & \underline{1.080} & 2.209 & 0.668 \\
\bottomrule
\end{tabular}}
\end{table}

\begin{table}[ht]
\centering
\caption{Per-dataset \rpet{} on the same seven datasets (lower is
better). Bold and underline mark the full-table best and
second-best values in each column.}
\vspace{1mm}
\label{tab:app_rpetrans}
\scriptsize
\setlength{\tabcolsep}{3.2pt}
\resizebox{\linewidth}{!}{%
\begin{tabular}{lrrrrrrr}
\toprule
\textbf{Method} & \textbf{7S} & \textbf{ETH} &
\textbf{KITTI} & \textbf{NRGBD} & \textbf{Oxford} &
\textbf{T\&T} & \textbf{VBR} \\
\midrule
VGGT-Omega~\citep{wang2026vggtomega} & 0.0172 & 0.101 & 2.358 &
0.0109 & 4.044 & 0.128 & 0.2557 \\
$\pi^3$~\citep{wang2025pi3} & 0.0158 & \underline{0.091} &
\underline{0.150} & 0.0091 & 1.999 &
\underline{0.106} & 0.1591 \\
GlueMap~\citep{pan2026gluemap} & \textbf{0.0145} & 0.583 &
1.805 & \underline{0.0079} & 2.268 & 0.115 & 8.599 \\
Scal3R~\citep{xie2026scal3r} & 0.0176 & 0.238 & 0.335 & 0.0092 &
\underline{0.259} & 0.107 & 0.2260 \\
VGGT-Long~\citep{deng2025vggtlong} & 0.0188 & 0.190 & 0.221 & 0.0098 & 0.997 &
0.110 & 0.2292 \\
VGGT-Align~\citep{zhang2026vggtalign} & 0.0188 & 0.191 & 14.88 & 0.0098 & 1.053 &
0.110 & 0.2633 \\
LingBot-MAP~\citep{chen2026lingbotmap} & 0.0205 & 0.495 & 6.004 & 0.0189 & 0.523 &
0.124 & 2.7208 \\
ZipMap~\citep{jin2026zipmap} & 0.0217 & 0.239 & 3.039 & 0.0160 & 3.005 &
0.144 & 0.1530 \\
VGGT~\citep{wang2025vggt} & 0.0221 & 0.244 & 4.708 & 0.0196 & 5.955 &
0.165 & 0.3632 \\
LoGeR~\citep{zhang2026loger} & 0.0223 & 0.289 & 4.110 & 0.0180 & 0.547 &
0.182 & 2.5563 \\
TTT3R~\citep{chen2026ttt3r} & 0.0263 & 1.229 & 0.329 & 0.0239 & 0.803 &
0.181 & 3.0620 \\
CUT3R~\citep{wang2025cut3r} & 0.0351 & 0.799 & 4.964 & 0.0397 & 1.054 &
0.172 & 3.1564 \\
\midrule
\textbf{\method{} (VGGT-Omega)} & \underline{0.0148} & \textbf{0.0567} &
\textbf{0.0931} & \textbf{0.0049} & \textbf{0.104} &
\textbf{0.103} & \textbf{0.0591} \\
\textbf{\method{} (LingBot-MAP)} & 0.0211 & 0.465 & 0.231 &
0.0196 & 0.285 & 0.147 & \underline{0.108} \\
\bottomrule
\end{tabular}}
\end{table}

\section{Per-dataset dense 3D geometry}
\label{app:e2e}

This appendix provides per-dataset surface reconstruction metrics for the dense geometry benchmark.
The dense-geometry evaluation set comprises $38$ sequences with dense 3D ground truth.
It includes $11$ sequences from ETH3D, $9$ from Neural RGB-D, and $18$ from 7-Scenes.
After Umeyama alignment, ICP uses a correspondence threshold of $0.25$\,m on ETH3D and $0.10$\,m on Neural RGB-D and 7-Scenes.
To weight each physical evaluation scene equally, Table~\ref{tab:e2e_omega} reports sequence-macro averages over these $38$ sequences.
Under unweighted dataset-macro averaging across the three datasets, VGGT-Omega baseline versus \method{} scores are $0.0796$ versus $0.0269$ for Chamfer distance, and $65.97$ versus $92.06$ for F1 score.
LingBot-MAP baseline versus \method{} scores are $0.1439$ versus $0.1014$ for Chamfer distance, and $61.34$ versus $74.58$ for F1 score.
Table~\ref{tab:step_geometry} isolates the relative geometric contributions of layout alignment and camera refinement.

\begin{table}[ht]
\centering
\caption{Dense 3D reconstruction metrics (Chamfer distance and F1 score) on VGGT-Omega versus \method{} evaluated on datasets with dense 3D ground truth.}
\vspace{1mm}
\label{tab:e2e_geom_omega}
\small
\setlength{\tabcolsep}{8.0pt}
\begin{tabular}{l cc cc}
\toprule
& \multicolumn{2}{c}{\textbf{Chamfer} $\downarrow$} &
\multicolumn{2}{c}{\textbf{F1} $\uparrow$} \\
\cmidrule(lr){2-3}\cmidrule(lr){4-5}
\textbf{Dataset} &
\textbf{Baseline} &
\textbf{\method{}} &
\textbf{Baseline} &
\textbf{\method{}} \\
\midrule
ETH3D & 0.0339 & 0.0334 & 98.92 & 99.00 \\
Neural RGB-D & 0.1310 & 0.0205 & 40.53 & 90.05 \\
7-Scenes & 0.0739 & 0.0269 & 58.45 & 87.13 \\
\bottomrule
\end{tabular}
\end{table}

\begin{table}[ht]
\centering
\caption{Dense 3D reconstruction metrics (Chamfer distance and F1 score) on LingBot-MAP versus \method{} evaluated on datasets with dense 3D ground truth.}
\vspace{1mm}
\label{tab:e2e_geom_map}
\small
\setlength{\tabcolsep}{8.0pt}
\begin{tabular}{l cc cc}
\toprule
& \multicolumn{2}{c}{\textbf{Chamfer} $\downarrow$} &
\multicolumn{2}{c}{\textbf{F1} $\uparrow$} \\
\cmidrule(lr){2-3}\cmidrule(lr){4-5}
\textbf{Dataset} &
\textbf{Baseline} &
\textbf{\method{}} &
\textbf{Baseline} &
\textbf{\method{}} \\
\midrule
ETH3D & 0.1295 & 0.1295 & 86.49 & 86.49 \\
Neural RGB-D & 0.1620 & 0.0675 & 42.66 & 65.24 \\
7-Scenes & 0.1403 & 0.1071 & 54.87 & 72.01 \\
\bottomrule
\end{tabular}
\end{table}

\section{Chunk and overlap experiments}
\label{app:windows}

Table~\ref{tab:windows} compares selected chunk and overlap sizes against fixed defaults in an exploratory sweep.
The default chunk size and overlap size are $60$ and $30$ frames for VGGT-Omega, and $120$ and $32$ frames for LingBot-MAP.
Each window configuration produces its own chunk predictions and optimized coarse layout, which is then held fixed during camera processing.
The coarse solve follows Appendix~\ref{app:protocol}, with a limit of $5000$ function evaluations.

Both window configurations use dense fine optimization for $T\le 2000$ and overlap consensus for longer sequences.
Overlap consensus averages rotations and translations at stride-$10$ frames, retains the final frame, and interpolates pose corrections to the remaining frames.
For long sequences, this replaces the stride-$5$ keyframe optimization used in the main benchmark.

On KITTI, LingBot-MAP achieves $3.508$\,m with overlap consensus, compared with $6.063$\,m for Full in Table~\ref{tab:app_ate}.
Selected versus default macro-averaged \ate{} across the seven datasets is $4.403$ versus $4.908$\,m for VGGT-Omega, and $3.822$ versus $3.825$\,m for LingBot-MAP.

\begin{table}[ht]
\centering
\caption{Chunk and overlap comparison. $C/O$ denotes chunk size and overlap size in frames. The ``vs'' columns report \ate{} with the indicated default sizes. Bold marks a change from Table~\ref{tab:hparams}.}
\vspace{1mm}
\label{tab:windows}
\small
\setlength{\tabcolsep}{3.0pt}
\begin{tabular}{llrclrc}
\toprule
\textbf{Dataset} &
\multicolumn{3}{c}{\textbf{VGGT-Omega}~\citep{wang2026vggtomega}} &
\multicolumn{3}{c}{\textbf{LingBot-MAP}~\citep{chen2026lingbotmap}} \\
\cmidrule(lr){2-4}\cmidrule(lr){5-7}
& $C{/}O$ & \ate{} & vs $60/30$
& $C{/}O$ & \ate{} & vs $120/32$ \\
\midrule
7-Scenes & $\mathbf{90/30}$ & $0.0366$ & $0.0381$
& $\mathbf{120/8}$ & $0.0801$ & $0.0793$ \\
ETH3D & $60/30$ & $0.0546$ & $0.0546$
& $120/32$ & $0.479$ & $0.479$ \\
KITTI & $\mathbf{60/16}$ & $9.352$ & $9.736$
& $120/32$ & $3.508$ & $3.508$ \\
NRGBD & $60/30$ & $0.0219$ & $0.0219$
& $\mathbf{160/32}$ & $0.0608$ & $0.0776$ \\
Oxford & $60/30$ & $1.579$ & $1.579$
& $120/32$ & $1.651$ & $1.651$ \\
T\&T & $60/30$ & $0.476$ & $0.476$
& $120/32$ & $0.587$ & $0.587$ \\
VBR & $\mathbf{120/30}$ & $19.30$ & $22.45$
& $120/32$ & $20.39$ & $20.39$ \\
\bottomrule
\end{tabular}
\end{table}

\section{Ablation on the coarse-only layout}
\label{app:coarse_stage}

Table~\ref{tab:coarse_long} compares camera accuracy after the coarse layout stage alone against the Full pipeline on three long sequences.
Shared-camera refinement lowers \rper{} and \rpet{} on all three sequences while leaving \ate{} close to the coarse-stage result.

\begin{table}[ht]
\centering
\caption{Trajectory accuracy of Layout Only versus Full on the same feed-forward predictions.}
\vspace{1mm}
\label{tab:coarse_long}
\small
\setlength{\tabcolsep}{3.2pt}
\begin{tabular}{lrcccccc}
\toprule
& &
\multicolumn{3}{c}{Layout Only} &
\multicolumn{3}{c}{Full} \\
\cmidrule(lr){3-5}\cmidrule(lr){6-8}
\textbf{Sequence} & $T$ &
\ate{} & \rper{} & \rpet{} &
\ate{} & \rper{} & \rpet{} \\
\midrule
KITTI seq00 & $4541$
& $9.750$ & $0.202$ & $0.131$
& $9.730$ & $0.154$ & $0.093$ \\
spagna\_train0 & $14{,}141$
& $10.256$ & $0.542$ & $0.048$
& $10.261$ & $0.535$ & $0.034$ \\
campus\_train1 & $11{,}671$
& $7.381$ & $0.230$ & $0.076$
& $7.387$ & $0.157$ & $0.045$ \\
\bottomrule
\end{tabular}
\end{table}

\section{Controlled layout-solver comparison}
\label{app:layout_controls}

The layout-solver study uses ten long sequences: KITTI sequence~00, all seven VBR sequences, and Droid-W downtown2 and downtown5~\citep{li2026droidw}.
All evaluated configurations share identical model predictions, candidate revisit pairs, and initialization.
Camera refinement and point corrections are disabled to isolate layout optimization.
The full coarse solve allows at most $5000$ function evaluations.
The five configurations evaluate baseline pairwise alignment, bounded point sampling, residual-fit weighting, progressive loop scheduling, and the complete optimization solver.
The final step combines adaptive continuation, Huber loss, and bounded restarts.
Appendix~\ref{app:protocol} specifies the continuation and restart rules and their evaluation budgets.

\begin{table}[ht]
\centering
\caption{Ablation of layout-solver components on extended sequences.
Total time includes measurement construction and all layout solves.
Final solve is the duration of the final least-squares call.}
\vspace{1mm}
\renewcommand{\arraystretch}{1.1}

\label{tab:layout_controls}
\small
\setlength{\tabcolsep}{4.8pt}
\begin{tabular}{l rrr rr}
\toprule
Configuration & \ate{} & \rper{} & \rpet{} & Graph Build \& Solve (s) & Final solve (s) \\
\midrule
\multicolumn{6}{l}{Clean loop measurements} \\
Baseline & 21.7751 & 1.31094 & 0.11296 & 12232.18 & 1404.54 \\
$+$ Bounded sampling & 21.7569 & 1.31093 & 0.11280 & 2004.20 & 1345.43 \\
$+$ Fit weighting & 21.7770 & 1.31084 & 0.11262 & 2077.49 & 1387.14 \\
$+$ Progressive loop scheduling & 21.5241 & 1.31077 & 0.11070 & 1477.59 & 688.02 \\
Complete & 17.4708 & 1.30972 & 0.09834 & 753.43 & 41.28 \\
\midrule
\multicolumn{6}{l}{Perturbed loop measurements} \\
Baseline & 20.4767 & 1.33444 & 0.11078 & 9343.00 & 1226.62 \\
$+$ Bounded sampling & 20.5269 & 1.33447 & 0.11109 & 1641.88 & 1220.56 \\
$+$ Fit weighting & 21.2226 & 1.33061 & 0.11627 & 1798.38 & 1284.72 \\
$+$ Progressive loop scheduling & 19.1915 & 1.32976 & 0.10568 & 1032.27 & 461.17 \\
Complete & 16.9298 & 1.32951 & 0.10280 & 794.13 & 31.31 \\
\bottomrule
\end{tabular}
\end{table}

We test solver robustness against noisy loop measurements by perturbing loop closures. Specifically, synthetic noise is injected into $25\%$ of detected loop edges using a fixed random seed.
For each selected edge, we apply a $10^\circ$ rotation error. We also inject a $25\%$ translation offset and a $0.10$ log-scale distortion.
The evaluation confirms efficiency gains from bounded sampling, alongside accuracy and runtime gains from the complete solver.
The complete configuration reduces clean mean \ate{} by $19.77\%$ relative to the baseline.
This improvement concentrates in VBR, where mean error falls from $28.6017$ to $22.4461$ m.
Peak host memory remains comparable across configurations.

\section{Placement and motion ablation settings}
\label{app:prediction_controls}

The four controls in Table~\ref{tab:prediction_controls} evaluate different strategies for resolving multiple camera predictions under identical optimized coarse layouts.
Layout Only selects the first available coarse-placed pose for each frame in chunk order.
Single selects the occurrence that maximizes the minimum distance, in frames, to either boundary of its chunk.
Ties are broken by the smaller chunk index.
Both Layout Only and Single directly output the selected coarse-placed poses without frame-level optimization.
Unary computes a robust pose average over all coarse-placed occurrences of each frame using deterministic Huber IRLS, without relative-motion factors.
Motion optimizes within-chunk relative-motion factors without placement priors.
It fixes the first camera to its coarse placement and uses $\lambda_f=1$ and $\lambda_p=0$.
This control allows up to $5000$ function evaluations with $\texttt{ftol}=\texttt{xtol}=\texttt{gtol}=10^{-8}$.
Shared jointly optimizes placement priors and relative-motion constraints as formulated in Eq.~\eqref{eq:fine_objective}, and corresponds to Full in the main benchmark.
All configurations use the same evaluation protocol, and results are macro-averaged across all seven datasets.

\section{Dense geometry across optimization stages}
\label{app:step_attribution}

Table~\ref{tab:step_geometry} extends the trajectory comparison in Table~\ref{tab:matched} to dense surface geometry on ETH3D, Neural RGB-D, and 7-Scenes.
Native sequentially stitches chunks through their overlaps without coarse or fine graph optimization.
Layout Only applies the optimized coarse chunk placements without frame-level corrections.
Full uses the same coarse layout and adds shared-camera corrections to the pointmaps, preserving their predicted depth and chunk scale.
Shared-camera refinement produces  additional gains in Chamfer and F1 on both backends.
Native sequentially stitches chunks through their overlaps without coarse or fine graph optimization.
Layout Only applies the optimized coarse chunk placements without frame-level corrections.
Full uses the same coarse layout and adds shared-camera corrections to the pointmaps, preserving their predicted depth and chunk scale.
Shared-camera refinement produces  additional gains in Chamfer and F1 on both backends.

\begin{table}[ht!]
\centering
\caption{Dense geometry across optimization stages. F1 is reported in percent.}
\label{tab:step_geometry}
\renewcommand{\arraystretch}{1.1}
\vspace{1mm}
\small
\begin{tabular}{@{}lrr@{\hspace{8mm}}lrr@{}}
\toprule
\multicolumn{3}{c}{VGGT-Omega} &
\multicolumn{3}{c}{LingBot-MAP} \\
\cmidrule(lr){1-3} \cmidrule(lr){4-6}
Output & Chamfer $\downarrow$ & F1 $\uparrow$ &
Output & Chamfer $\downarrow$ & F1 $\uparrow$ \\
\midrule
Native      & 0.0759 & 65.92 & Native      & 0.1423 & 61.13 \\
Layout Only & 0.0273 & 91.20 & Layout Only & 0.1049 & 74.49 \\
Full        & 0.0272 & 91.26 & Full        & 0.1042 & 74.60 \\
\bottomrule
\end{tabular}
\end{table}

\end{document}